\documentclass[lettersize,journal]{IEEEtran}

\usepackage[nolist, nohyperlinks]{acronym}
\usepackage{amsmath}
\usepackage{array}
\usepackage[style=ieee,hyperref,backref,natbib,sorting=nyt,sortcites,maxbibnames=8,mincitenames=1,maxcitenames=2,backend=bibtex]{biblatex}
\usepackage{booktabs}
\usepackage{catchfile}
\usepackage{makecell}
\usepackage{subcaption}
\usepackage{tikz}
\usepackage{todonotes}
\usepackage{rotating} 

\begin{acronym}
	\acro{AFD}{Active Fire Detection Index for Sentinel 2}
	\acro{ASIC}{Application-Specific Integrated Circuit}
	\acro{BOA}{Bottom of Atmosphere}
	\acro{CNN}{Convolutional Neural Network}
	\acro{CPU}{Central Processing Unit}
	\acro{DAG}{Directed Acyclic Graph}
	\acro{ESR}{Expected Scalarized Return}
	\acro{FPU}{Floating Point Unit}
	\acro{FLOP}{Floating Point Operation}
	\acro{GCN}{Graph Convolutional Network}
	\acro{GNN}{Graph Neural Networks}
	\acro{GP}{Gaussian Process}
	\acro{GPU}{Graphics Processing Unit}
	\acro{IR}{Infrared}
	\acro{LASSO}{Least Absolute Shrinkage and Selection Operator}
	\acro{LEO}{Low Earth Orbit}
	\acro{LSTM}{Long Short-Term Memory}
	\acro{MLP}{Multi-Layer Perceptron}
	\acro{NAS}{Neural Architecture Search}
	\acro{NBR}{Normalized Burn Ratio}
	\acro{NDVI}{Normalized Difference Vegetation Index}
	\acro{OLS}{Ordinary Least Squares}
	\acro{OSM}{Open Street Maps}
	\acro{RMSE}{Root Mean Squared Error}
	\acro{SER}{Scalarized Expected Return}
	\acro{SGD}{Stochastic Gradient Descent}
	\acro{SVM}{Support Vector Machine}
	\acro{TPU}{Tensor Processing Unit}
	\acro{TOA}{Top of Atmosphere}
	\acro{UAS}{Unmanned Aircraft Systems}
	\acro{USB}{Universal Serial Bus}
\end{acronym}

\DeclareUnicodeCharacter{03C6}{\ensuremath{\phi}}
\newcolumntype{?}{!{\vrule width 1pt}}

\title{Designing a Classifier for Active Fire Detection from Multispectral Satellite Imagery Using Neural Architecture Search}
\author{Amber Cassimon, Phil Reiter, Siegfried Mercelis, Kevin Mets \thanks{A. Cassimon, P. Reiter, S. Mercelis and K. Mets are with IDLab - Faculty of Applied Engineering, University of Antwerp - imec}}
\date{\today}

\newcommand\submittedtext{%
  \footnotesize This work has been submitted to the IEEE for possible publication. Copyright may be transferred without notice, after which this version may no longer be accessible.}

\newcommand\submittednotice{%
\begin{tikzpicture}[remember picture,overlay]
\node[anchor=south,yshift=10pt] at (current page.south) {\fbox{\parbox{\dimexpr0.65\textwidth-\fboxsep-\fboxrule\relax}{\submittedtext}}};
\end{tikzpicture}%
}

\begin{document}

\maketitle

\begin{abstract}
	This paper showcases the use of a reinforcement learning-based \ac{NAS} agent to design a small neural network to perform active fire detection on multispectral satellite imagery.
	Specifically, we aim to design a neural network that can determine if a single multispectral pixel is a part of a fire, and do so within the constraints of \iac{LEO} nanosatellite with a limited power budget, to facilitate on-board processing of sensor data.
	In order to use reinforcement learning, a reward function is needed.
	We supply this reward function in the shape of a regression model that predicts the F1 score obtained by a particular architecture, following quantization to INT8 precision, from purely architectural features.
	This model is trained by collecting a random sample of neural network architectures, training these architectures, and collecting their classification performance statistics.
	Besides the F1 score, we also include the total number of trainable parameters in our reward function to limit the size of the designed model and ensure it fits within the resource constraints imposed by nanosatellite platforms.
	Finally, we deployed the best neural network to the Google Coral Micro Dev Board and evaluated its inference latency and power consumption.
	This neural network consists of 1,716 trainable parameters, takes on average 984{\textmu}s to inference, and consumes around 800mW to perform inference.
	These results show that our reinforcement learning-based \ac{NAS} approach can be successfully applied to novel problems not tackled before.
\end{abstract}

\begin{IEEEkeywords}
Neural Architecture Search, Deep Learning, AutoML, Multispectral Imaging, Active Fire Detection
\end{IEEEkeywords}

\section{Introduction}
\label{sec:intro}
	\IEEEPARstart{M}{ultispectral} satellite data has many uses ranging from estimating heat storage in urban areas~\cite{Hrisko_2021_Estimating}, bathymetry~\cite{Tonion_2020_AMachine} or monitoring the evolution of rivers~\cite{Cavallo_2022_Monitoring}.
	Analyzing large volumes of data by hand is cumbersome, however.
	Because of this, deep learning techniques have been successfully introduced to automate the analysis of multispectral satellite imagery~\cite{Barros_2022_Multispectral,Vali_2020_Deep}.
	While accurate analyses can be made using deep learning, these neural networks often come with high computational costs.
	This makes deep learning approaches infeasible in environments and tasks where computational resources and power are at a premium, such as when performing processing of multispectral satellite imagery on-board smallsats~\cite{Daghouri_2023_Investigating,Dahbi_2017_Power}.
	\textcite{Daghouri_2023_Investigating} size the electrical power system for a nano satellite at several watts, ranging from around 6W per solar panel at peak times, down to just 1.7W during low times.
	\textcite{Dahbi_2017_Power} arrive at a similar number, with their nanosatellite generating between 0W and 3.5W depending on the precise position of the satellite in an orbit.
	Deep learning systems can be designed to operate in such low-power environments, but this process is often complex and laborious, requiring experienced engineers to iteratively design neural networks that maximally take advantage of the available resources~\cite{Giuffrida_2022_ThePhi}.
	In this paper, we present \iac{NAS} system that is capable of automatically designing neural networks for the task of active fire detection from multispectral satellite imagery considering both the designed networks' task-performance (measured by the F1 score they achieve) as well as the required computational resources (using the total number of trainable parameters as a proxy for resource consumption).
	We deploy a neural network onto a Google Coral Micro development board~\cite{Google_2022_Dev}, measuring the power requirements of the system to ensure it can reasonably fit within the power envelope afforded by smallsat earth observation missions.
	The remainder of this paper is structured as follows.
	In Section~\ref{sec:related_work} we consider state-of-the-art research in various fields related to our use-case.
	Section~\ref{sec:methods} details how the \ac{NAS} agent was designed, and how we achieved the prerequisites.
	Next, Section~\ref{sec:experiments} discusses the experiments we performed, listing experimental setup and used parameters in detail.
	Finally, we provide some closing thoughts on our system in Section~\ref{sec:discussion}.

\submittednotice

\section{Related Work}
\label{sec:related_work}
	This section will discuss existing approaches in the field of reinforcement learning-based \ac{NAS} (Section~\ref{subsec:related_work:nas}), multispectral image processing (Section~\ref{subsec:related_work:multispectral_image_processing}) and active fire detection (Section~\ref{subsec:related_work:active_fire_detection}).

	\subsection{\acl{NAS}}
	\label{subsec:related_work:nas}
		\Ac{NAS} has been used to design neural networks that outperform human-designed neural networks in a wide variety of domains, including computer vision~\cite{Real_2017_Large}, natural language generation~\cite{Javaheripi_2022_LiteTransformerSearch}, and wind forecasting~\cite{Pujari_2023_Better}.
		The variety of techniques that have been used in \ac{NAS} is almost as wide as the set of problem domains that have been tried, including Bayesian approaches~\cite{Kandasamy_2018_Neural}, evolutionary algorithms~\cite{Elsken_2019_Efficient,Real_2017_Large}, continuous relaxation~\cite{Liu_2019_Darts}, graph diffusion~\cite{Asthana_2024_Multiconditioned} and reinforcement learning~\cite{Pham_2018_Efficient,Li_2023_GraphPNAS}.
		Despite having fallen out of use in recent years, recent innovations in the space of reinforcement learning-based \ac{NAS} approaches encouraged us to use reinforcement learning in this work.
		Since this paper uses a reinforcement learning-based approach, we will focus on reinforcement learning for the remainder of this section.

		Some of the first work in the field of \ac{NAS} was done using reinforcement learning~\cite{Zoph_2017_Neural,Pham_2018_Efficient}.
		\textcite{Zoph_2017_Neural,Pham_2018_Efficient} used \iac{LSTM}-based reinforcement learning agent to sequentially sample architectural decisions and build the computational graph of the neural network.
		They evaluated their approach in both the language generation and computer vision domains, and found their approach surpassed human-designed neural networks, achieving strong performance on both domains.
		\textcite{Pham_2018_Efficient} considered both a macro and micro search space.
		In the macro search space, their reinforcement learning agent designed the entire neural network, while in the micro search space, it designed a small cell that was repeated multiple times to form a complete neural network.

		\textcite{Cassimon_2024_Scalable} introduce a novel reinforcement learning agent that iteratively improves on a given neural network architecture by making small alterations to the architecture.
		Their transformer-based reinforcement learning agent is evaluated on two \ac{NAS} benchmarks focused on computer vision applications: NAS-Bench-101~\cite{Ying_2019_NASBench101} and NAS-Bench-301~\cite{Siems_2021_Nasbench}.
		They find their agent is capable of finding strong architectures on both benchmarks and scales well with the size of the search space.
		Contrary to this work, \citeauthor{Cassimon_2024_Scalable} make use of a look-up table and a pre-trained gradient boosted tree model as their reward function for the NAS-Bench-101 and NAS-Bench-301 benchmarks, respectively.

		\Ac{NAS} methods require a method to find the performance of a specific neural network architecture on a particular task.
		The most naive way of achieving this is by simply training neural networks to convergence, as was done in early \ac{NAS} works~\cite{Zoph_2017_Neural}.
		In recent literature, this is usually achieved through performance prediction methods: Methods designed to predict the performance a specific neural network on a specific task, assuming a particular fixed training procedure and set of hyperparameters.
		Some methods make use of a regression model that takes architectural features as input and outputs a prediction for the neural network's target performance~\cite{Lu_2023_PINAT}.
		Others use what are called zero-cost proxies: Methods that don't require any training (Of the designed network or the performance predictor), but usually do require inference~\cite{Kadlecova_2024_Surprisingly,Mellor_2021_Neural}.
		There also exist one-shot methods that rely on concepts like weight sharing~\cite{Pham_2018_Efficient}.

		\textcite{Lu_2023_PINAT} propose a transformer-like \ac{NAS} performance predictor which uses permutation-invariance modules to improve predictor performance in the face of graph isomorphism between different representations of the same architecture.
		They test their method on several existing benchmarks including NAS-Bench-101~\cite{Ying_2019_NASBench101} and NAS-Bench-201~\cite{Dong_2020_NAS}.
		Their PINAT method is also evaluated on the DARTS~\cite{Liu_2019_Darts} and ProxylessNAS~\cite{Cai_2018_Proxylessnas} search spaces.
		The architectures designed by PINAT achieve performance rivalling that of other state-of-the-art methods.

	\subsection{Multispectral Image Processing}
	\label{subsec:related_work:multispectral_image_processing}
		Multispectral satellite imagery can be used for a broad variety of use-cases.
		In this section, we examine some state-of-the-art deep-learning based approaches for analyzing multispectral imagery.

		\textcite{Zheng_2020_FPGA} present a framework for patch-free global learning of hyperspectral imagery.
		They achieve this using an encoder-decoder architecture enhanced using lateral connections.
		The encoder maps the entire image into a latent space, while the decoder decodes the latent image representation into a per-pixel classification map.
		Their method is supported by a global stochastic stratified sampling strategy ($\text{GS}^{2}$) to ensure a diversity of gradients and prevent convergence issues due to a low amount of training samples.
		Because hyperspectral datasets like Pavia University~\cite{Gamba_2013_Pavia} only contain a single image, training must also occur in batches with a batch size of 1.
		This can lead to issues with batch normalization operations included in many models, thus the authors opt to replace batch normalization with group normalization~\cite{Wu_2018_Group}.
		They also introduce a spectral attention module to reweight the feature maps of their hyperspectral images.
		Through these techniques, they obtain strong results on the Pavia University~\cite{Gamba_2013_Pavia}, Salinas~\cite{Salinas_2019_Salinas} and CASI University of Houston datasets~\cite{Hyperspectral_2018_Houston}.
		The paper also demonstrates the computational efficiency of their approach by a comparison in terms of the number of \acp{FLOP} and the number of trainable parameters.

		\textcite{Kemker_2018_Algorithms} use existing computer vision algorithms~\cite{Lin_2017_RefineNet,Pinheiro_2016_Learning} to perform semantic segmentation on multispectral imagery.
		To address the lack of labelled data, they make use of a synthetic dataset generated using a simulator.
		They also introduce a new high-resolution multispectral dataset captured from \iac{UAS}, RIT-18, and evaluate several computer vision approaches on the new dataset.
		Their results show that end-to-end \ac{CNN} based segmentation models can outperform classical classification approaches including k nearest neighbours, \ac{SVM}, \ac{MLP}, spatial mean pooling and unsupervised learning approaches such as MICA~\cite{Kemker_2017_Self} and SCAE~\cite{Kemker_2017_Self}.
		They also conclude that the use of synthetically generated data to initialize models can improve performance over randomly initialized models.

	\subsection{Active Fire Detection}
	\label{subsec:related_work:active_fire_detection}
		In this section, we will consider the state-of-the-art in active fire detection from multispectral satellite imagery.
		Research exists into active fire detection methods for terrestrial purposes~\cite{Gunay_2010_Fire}, this paper focuses on the detection of fires from satellite imagery.

		\textcite{Barmpoutis_2020_AReview} provide an overview of methods for early fire detection based on optical remote sensing.
		They categorize methods in terms of where the sensors are placed: terrestrial, airborne or space borne.
		They consider optical remote sensing systems operating in the visible and \ac{IR} spectrum, as well as multispectral systems based on traditional machine learning and deep learning.
		In their conclusion, \textcite{Barmpoutis_2020_AReview} highlight smallsat-based fire detection system as a promising avenue for improving the detection of active wildfires.

		\textcite{Florath_2022_Supervised} build a system to detect active fires and burnt areas simultaneously based on supervised machine learning.
		They tackle several challenges including the generation of good reference data, the detection of active fires and burned areas at a high spatial resolution, while at the same time trying to keep their methodology as generic as possible.
		To generate useful data for their machine learning models, they leverage a combination of \ac{OSM} data, vector data from governmental agencies and Sentinel 2 L2A \ac{BOA} products.
		\textcite{Florath_2022_Supervised} evaluate seven different supervised machine learning models, including gradient boosting, extremely randomized trees, \acp{MLP} and \acp{CNN}.
		They conclude that the performance of all models is satisfactory when it comes to detecting fire, but find most models struggle more with the classification of burned areas.
		They hypothesize that the models struggle with the separation of burned and unburned areas because of their spectral similarity.

		To gather sufficient labelled data to train a deep learning system for active fire detection, we use the algorithm that \textcite{Massimetti_2020_Volcanic} used to detect and monitor volcanic activity.
		\textcite{Meoni_2023_THRawS} found this algorithm useful for detecting fire events on Sentinel 2 data.
		The algorithm proposed by \textcite{Massimetti_2020_Volcanic} is relatively basic.
		It consists of the computation of reflectance ratios between different bands of Sentinel 2 data.
		These reflectance ratios are then compared against reference thresholds and the result of these comparisons is combined through a series of distinct logical tests.
		Four different logical tests are used, and if at least one of them returns true, a pixel is flagged as a hot spot.

		\textcite{Xu_2023_Sentinel} discuss the challenges faced in developing active fire detection products from Sentinel 3 data.
		Sentinel 3 carries various instruments and can capture information in \ac{IR} channels  with a spatial resolution of 1km~\cite{Coppo_2010_SLSTR}.
		They form a daytime active fire detection product by combining different spectral bands and adapting existing active fire detection methods designed for nighttime detection of active fires based on a single \ac{IR} band~\cite{Xu_2020_First}.
		The original active fire detection algorithm relies on a series of masking operations using varying thresholds, atmospheric corrections and cluster detections.
		The new active fire detection product is compared against existing products and found to have similar performance, with nuanced differences in which fires are detected by both products.


\section{Methods}
\label{sec:methods}
	This section will describe the various aspects of our research in more detail, including the target task considered (Section~\ref{subsec:methods:target_task}), the way we gathered data on the training performance of our neural networks (Section~\ref{subsec:methods:performance_data_gathering}), the training procedure for our performance predictors (Section~\ref{subsec:methods:performance_predictor_training}) and the training procedure for our \ac{NAS} agent (Section~\ref{subsec:methods:nas_agent_training}).

	\subsection{Target Task}
	\label{subsec:methods:target_task}
		\begin{figure*}
			\centering
			\includegraphics[width=\textwidth, clip]{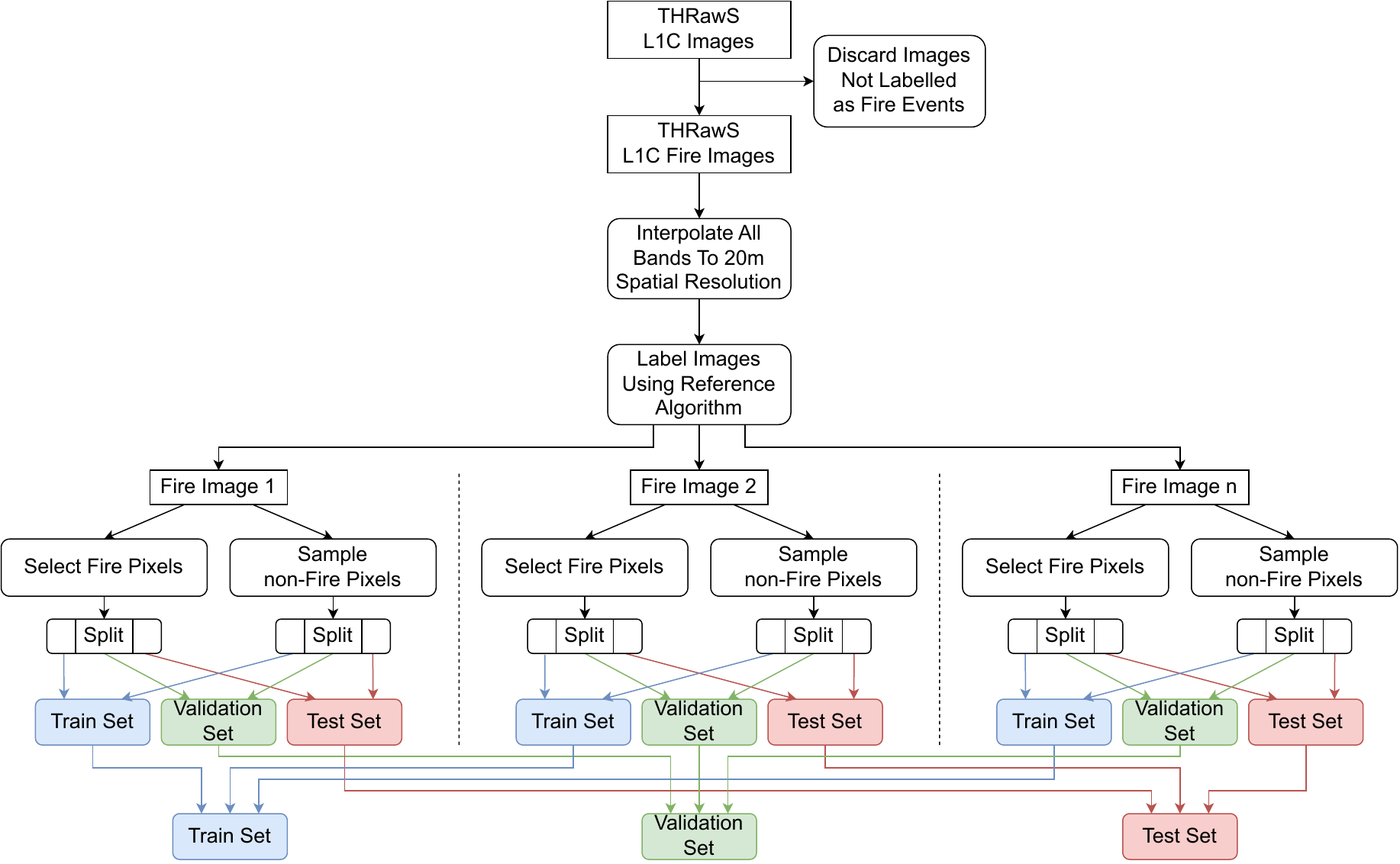}
			\caption{A schematic overview of the procedure used to generate the dataset.}
			\label{fig:methods:dataset_selection_procedure}
		\end{figure*}

		The task considered in this paper is the detection of active wildfires from Sentinel-2 multispectral satellite imagery.
		The dataset used was a subset of the THRawS dataset~\cite{Meoni_2023_THRawS}.
		Specifically, we selected all 20 tiles that contain ``fire'' events at the time of writing.
		We use L1C \ac{TOA} products, since they do not require the computationally expensive atmospheric correction process necessary for L2A \ac{BOA} products~\cite{Meoni_2023_THRawS}.
		Wildfires will be detected on a per-pixel basis, i.e., we will consider the spectral data for a single pixel at a time (13 bands), and output a binary classification: a pixel is labelled as either ``fire'' or ``no fire''.
		Following~\textcite{Meoni_2023_THRawS}, we use the algorithm from~\textcite{Massimetti_2020_Volcanic} to obtain a binary classification mask for the entire image, specifically, we use the implementation provided by PyRawS~\cite{Del_Prete_2023_THRawS}.
		Given that the different spectral bands in a Sentinel 2 image have different spatial resolutions, we resample all bands to the 20m resolution.
		This follows the 20m spatial resolution used by~\textcite{Massimetti_2020_Volcanic} in their hotspot detection algorithm.
		In the THRawS satellite images, negative label (``no fire'') pixels significantly outnumber positive label pixels.
		To compensate for this, we first select all fire pixels, and then randomly sample an equal number of ``no fire'' pixels at random from each satellite image.
		In total, we find $110,448$ positive label pixels, and sample an equal amount of negative label pixels, leading to a total of $220,896$ samples in the complete dataset.
		This is done on a per-image basis, to ensure that there is no distribution shift in regard to geographic regions, vegetation, etc. between both the positive and negative pixels.
		Following this, in each image individually, positive and negative pixels are split in a 70\%/15\%/15\% fashion to create a training, validation and test subset.
		Thus, the training set contains $154,636$ samples, while the validation and test sets each contain $33,130$ samples.
		The individual training, validation and test subsets of all images are then merged into datasets that span all images.
		This procedure is illustrated in Figure~\ref{fig:methods:dataset_selection_procedure}.

		\begin{figure}
			\centering
			\includegraphics[width=\columnwidth, clip]{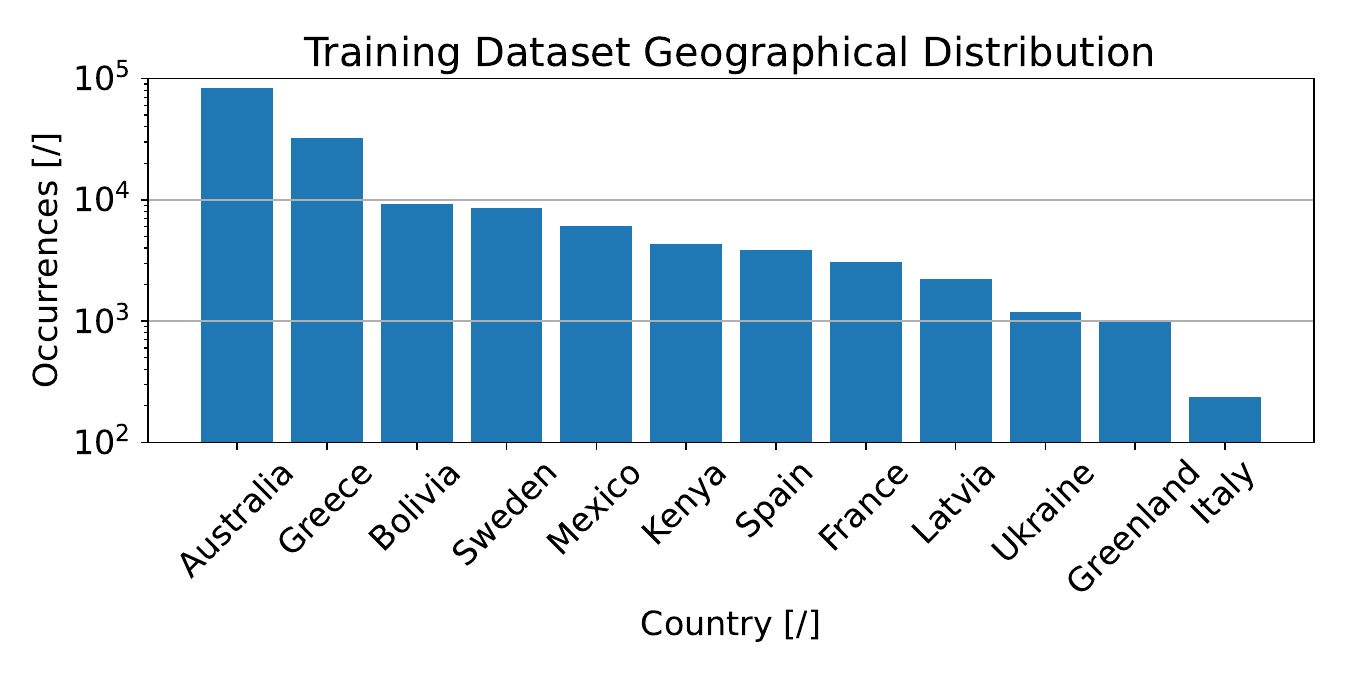}
			\caption{A histogram of the geographic distribution of our dataset. Note the logarithmic Y-axis.}
			\label{fig:methods:geographic_distribution}
		\end{figure}

		Figure~\ref{fig:methods:geographic_distribution} shows the geographic distribution of our data points.
		This distribution is identical between the training, validation and test set.
		Our data is clearly not distributed uniformly in terms of geography, with more than half of all samples in the training set ($82,938$ or $53.63\%$) originating from Australia, while only $236$ our of $154,636$ samples ($0.15\%$) originate in Italy.

		To aid our classification algorithm, we include three indices as a form of feature engineering.
		Specifically, we compute the \ac{NBR}~\cite{Key_2006_Landscape}, \ac{NDVI}~\cite{Pettorelli_2013_TheNormalized} and \ac{AFD}~\cite{Cicala_2018_Landsat}, and append these as features to our pixel data to improve classification performance, giving each pixel a total of 16 features (13 spectral bands and 3 indices).
		Since our goal is to eventually deploy the designed neural networks to an embedded device, we include the computation of these indices as part of the model.
		This ensures that the device can accept data without requiring that preprocessing is done externally.

	\subsection{Performance Data Gathering}
	\label{subsec:methods:performance_data_gathering}
		\ac{NAS} algorithms require a way to evaluate the performance of a neural network, preferably without training it to convergence.
		While many approaches have been proposed over the years, including one-shot methods~\cite{Pham_2018_Efficient} and zero-cost proxies~\cite{Kadlecova_2024_Surprisingly}, in the context of this work, we opted for a performance prediction model~\cite{White_2021_How}.
		To train such a model, we require performance data from trained neural networks.
		Often, gathering large amounts of such training data is computationally expensive due to the computational requirements of a single neural network.
		For our work however, the computational resources needed for training a single neural network are very limited, given that the network must be able to operate in a power envelope of several watts.
		This allows us to more easily gather performance data by training many neural networks.
		Alternative approaches such as one-shot methods exist, but one-shot methods are complex to train correctly, and if done incorrectly, yield low ranking correlation with the ground-truth data~\cite{Yu_2020_How}.
		Zero-cost proxies are another alternative, but they often require instantiating the designed neural network and performing inference on it~\cite{Mellor_2021_Neural}.
		These limitations make performance prediction models an attractive alternative to one-shot methods and zero-cost proxies in our case.
		We trained as many neural networks as possible within a fixed computational budget of 14 days (2 weeks) using 12 \acp{CPU} and a single NVIDIA A100 80GB PCIe \ac{GPU}.
		The limited size of the neural networks being trained actually makes it feasible to train multiple neural networks concurrently on a single \ac{GPU}.
		Exact parameters of the training process can be found in Section~\ref{subsec:experiments:perf_data_anal}.

		Training was done using version 2.15 of the TensorFlow framework~\cite{Abadi_2015_TensorFlow}, given that the use of TensorFlow Lite would be required to deploy the models later on.
		Following training, neural networks were quantized to INT8 precision using post-training quantization, with the aim of eventually deploying them on the Google Coral Micro Dev Board.
		Sometimes, quantization failed, with TensorFlow Lite reporting a violation of the same scale constraint.
		All inputs to a concatenation operation in TensorFlow Lite are required to use the same scale and zero point quantization parameters, and TensorFlow Lite's quantization systems failed to satisfy this constraint in a number of cases.
		Usually, this was the consequence of the computation of the different indices (\ac{NBR}, \ac{NDVI} and \ac{AFD}) being folded into the model.
		TensorFlow Lite failed to quantize the division operations used in these indices, which resulted in quantization and dequantization operations being inserted around them.
		The additional quantization and dequantization operations resulted in different quantization parameters being used when the different indices were concatenated with the original features they were calculated from.
		When this happened, it was impossible for us to ascertain the post-quantization F1 score of the trained neural network, thus, this instance of the architecture was ignored.
		Other training runs with different random initializations for the same architecture were included in the dataset.

		Neural networks were randomly sampled from the search space using the same algorithm that \textcite{Cassimon_2024_Scalable} use to sample initial states for their reinforcement learning environment.
		Our search space is a macro search space, that requires selecting the topology and node labels of a computational graph, similar to the cells considered by \textcite{Ying_2019_NASBench101}.
		We consider architectures with up to 8 nodes including one input and output node, leaving 6 nodes requiring operation labels to be assigned by the \ac{NAS} agent.
		Node labels are selected from a set of 10 possible labels: ``linear-prelu'', ``linear-relu'', ``linear-relu6'', ``linear-tanh'', ``linear'', ``conv-3'', ``conv-5'', ``max-pool-3'', ``max-pool-5'' and ``spectral-attn''.
		Preliminary experiments showed that linear classifiers can perform fairly well with the features we use, thus, we opted to include the ``linear'' operation, which is a simple linear transformation without non-linearity, to offer the \ac{NAS} agent the freedom to design both linear and non-linear classifiers.
		We operate on a single pixel at a time, rather than on a patch of multiple pixels as is often the case.
		Given that we have a single flat feature vector, we opted to include a linear layer with a variety of non-linearities in the form of the ``linear-prelu'', ``linear-relu'', ``linear-relu6'' and ``linear-tanh'' operations.
		Since spectral features in our dataset are roughly ordered by frequency, the first 13 features do include a notion of locality.
		To exploit this locality and allow the exploitation of relations between neighbouring frequency bands, we included both 1-D convolution and pooling operations in the search space through the ``conv-3'' and ``conv-5'', ``max-pool-3'' and ``max-pool-5'' operations.
		All convolution and pooling operations use ``same'' padding and operate with a stride of 1.
		Finally, we also include the spectral attention operation first proposed by \textcite{Zheng_2020_FPGA}.
		From this information, we can compute an upper bound on the size of the search space, $\Omega$.
		Our neural networks have between 2 and 8 vertices ($v \in \left[2, 8\right]$).
		For each vertex count, we need to label $v - 2$ nodes with one of 10 ($\left|o\right| = 10$) operations.
		$v$ Vertices in \iac{DAG} can be connected with between $v - 1$ and $\frac{v \cdot \left(v - 1\right)}{2}$ edges ($e \in \left[v - 1, \frac{v \cdot \left(v - 1\right)}{2}\right]$).
		The number of ways we can sample $d$ edges from a set of $e$ possible edges can be expressed mathematically as a combination.
		Combined, this leads to the expression in equation~\ref{eq:search_space_size_upper_bound} as an upper bound on the size of the search space.
		In reality, the number of unique architectures in the search space will be lower, since not every set of edges is a valid set of edges, and isomorphism hasn't been accounted for.
		Still, this number can give us an idea of the size of the search space we are operating in.

		\begin{equation}
		\label{eq:search_space_size_upper_bound}
			\left|\Omega\right| \leq \sum_{v=2}^{8}{\sum_{e=v - 1}^{\frac{v \cdot \left(v - 1\right)}{2}}{{\frac{v \cdot \left(v - 1\right)}{2} \choose e} \cdot \left(v - 2\right)^{|o|}}}
		\end{equation}

		Evaluating the expression in Equation~\ref{eq:search_space_size_upper_bound} for our case yields $268,143,512,722,241$ or $2.68 \times 10^{14}$, roughly 4 orders of magnitude smaller than the DARTS search space~\cite{Liu_2019_Darts} for combined normal and reduction cells, but 9 orders of magnitude larger than the NAS-Bench-101 search space~\cite{Ying_2019_NASBench101} for a single cell.

		\begin{figure}
			\centering
			\includegraphics[width=.75\columnwidth, clip]{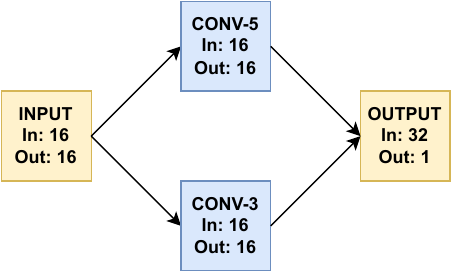}
			\caption{An example network involving multiple inputs to the output node.}
			\label{fig:ruleset:example4}
		\end{figure}

		\begin{figure}
			\centering
			\includegraphics[width=\columnwidth, clip]{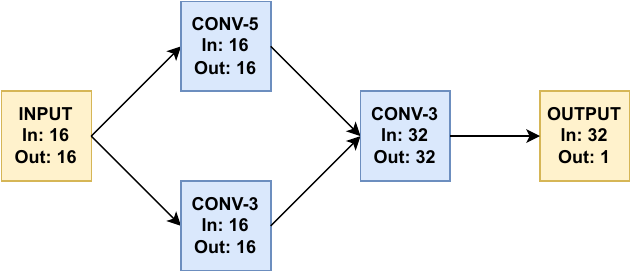}
			\caption{An example network involving multiple inputs to an intermediate node.}
			\label{fig:ruleset:example5}
		\end{figure}


		In order to go from a neural network architecture expressed as a computational \ac{DAG} to an actual neural network, a ruleset is needed to compute things like the number of filter channels in a convolutional neural network or the number of hidden units in recurrent neural network.
		We used the following rules to determine the number of features a node outputs:
		\begin{enumerate}
			\item Input nodes are a no-op, and thus output the 16 features they are given.
			\item Output nodes concatenate all inputs and sum them together to obtain the final probability of a pixel containing an active fire.
			\item Intermediate nodes concatenate all inputs before running them through their respective operations.
		\end{enumerate}

		Figures~\ref{fig:ruleset:example4} and~\ref{fig:ruleset:example5} demonstrate how the number of features that each operation outputs is determined.

	\subsection{Performance Predictor Training}
	\label{subsec:methods:performance_predictor_training}
		As mentioned at the start of Section~\ref{subsec:methods:performance_data_gathering}, \ac{NAS} methods require a method to ascertain the performance of a given neural network on a given task, without training the neural network to performance.
		We opted to train a supervised learning model to predict this performance from architectural features.
		Reinforcement learning requires sampling many (on the order of $10^{4}$ -- $10^{7}$, depending on the algorithm) transitions, and thus also a high number of evaluations of the reward function.
		From this, it follows that a reward function must be quick to evaluate, on the order of milliseconds per execution.
		Currently, the best way to meet this performance requirement is through the use of a small machine learning model to predict the performance of an architecture once trained.
	While alternatives exist, such as one-shot methods~\cite{Li_2019_Random,Liu_2019_Darts} and zero-cost proxies~\cite{Kadlecova_2024_Surprisingly,Mellor_2021_Neural}, many are still prohibitively expensive to calculate.
		Since we will be deploying the designed neural networks to an embedded device that requires INT8 quantization, we should also evaluate the post-quantization performance of the model.
		More specifically, we predict the model's post-quantization F1 score based on architectural features.
		We use the upper triangle of the adjacency matrix, combined with a one-hot encoded set of operations and the number of vertices and edges (as integers) as features for our predictor.

		To find a good prediction model, we consider a wide variety of linear and non-linear regression models.
		We use 4 different training strategies for our linear regression model: \ac{OLS}, Ridge, \ac{LASSO} and \ac{SGD}.
		We also consider 8 non-linear regression models with varying degrees of complexity: \acp{GP}, Random Forests, \acp{SVM}, K-Nearest Neighbour Regression, Radius Neighbour Regression, Gradient Boosted Trees, Multi-Layer Perceptrons and \acp{GNN}.
		Finally, we also include two random sampling methods to contextualize our results: One method that samples uniformly, and one method that samples according to a normal distribution with the mean and standard deviation matching that of the training set.

		All models except the \ac{GNN} and both random samplers are taken from the scikit-learn Python library~\cite{Pedregosa_2011_Scikit}.
		Models were trained using 5-fold cross validation with a held-out test set to be used later for evaluation.
		We consider 4 different evaluation metrics: the Pearson correlation coefficient, Kendall's $\tau$, the coefficient of determination ($R^{2}$) and the \ac{RMSE} obtained on the training and validation set.

	\subsection{NAS Agent Training}
	\label{subsec:methods:nas_agent_training}
		Once a sufficiently accurate performance prediction model has been trained, we can start building \iac{NAS} agent.
		In this setting, the performance prediction model is used as the reward function for the reinforcement learning agent.

		Since we are considering the case of on-board processing, we must also consider the computational resources required to evaluate the designed neural networks.
		A large and complex neural network might provide accurate predictions, but if we are unable to execute the network on-board the satellite because of its computational requirements, it is of limited use.
		There are a number of possible metrics that can be used for this.
		Some parameters can easily be computed from the neural network architecture itself, and are quick to evaluate.
		This includes parameters like the number of \acp{FLOP} required to evaluate a neural network, the total number of trainable parameters, the total working set size~\cite{Liberis_2021_uNAS}, etc.
		Such parameters have the advantage that they are independent of the device the neural network gets deployed on, thus allowing the same \ac{NAS} agent to be used to design neural networks for different hardware devices without invalidating the results.

		It is also possible to instantiate the neural network, execute it on a device, and measure certain parameters like the inference latency or the energy consumption.
		Compared to using a basic parameter like \ac{FLOP} count, measuring these metrics produces a much more accurate metric of what the actual impact is of certain design decisions.
		The disadvantage of using a measured metric is that it must be measured, which can be a slow and cumbersome process.
		Updating the neural network used on a microcontroller usually requires re-writing some form of permanent storage such as on-board flash memory or an SD card, which can be slow.
		Similar to the performance prediction model used earlier, our method of evaluating the resource requirements of a neural network must be executable in milliseconds to be able to gather the required number of samples to train an effective reinforcement learning agent.
		This makes the use of measured metrics infeasible in our setting.

		Recently, there has also been research into predicting measured metrics such as latency and energy consumption~\cite{Kadlecova_2024_Surprisingly,Yash_2024_On}.
		Such prediction models provide a viable alternative to measuring these metrics in the context of the constraints imposed by our reinforcement learning setting.

		However, when a performance prediction model is combined with a separate model for predicting energy consumption, the combined computational cost still exceeds what is acceptable in a reinforcement learning setting.
		A combined machine learning model that predicts both energy consumption and task-performance may be able a viable option, but is considered outside of the scope for this publication.
		Thus, in this paper, we use the total number of trainable parameters as a proxy for the resource consumption of the designed neural network on the target device.

		So far, we have discussed the individual optimization objectives considered in this paper, but we have yet to touch on the subject of how multi-objective reinforcement learning is applied to solve this optimization problem.
		Regardless of the exact setting, optimizing for multiple conflicting objectives simultaneously always involves the selection of trade-offs between the advantages and disadvantages of different solutions to a problem.
		In the multi-objective reinforcement learning literature, one approach to resolving such trade-offs is the use of a utility-based approach first proposed by \textcite{Roijers_2013_ASurvey}.
		The utility-based approach requires the definition of a utility function that captures how a user derives utility from a reinforcement learning policy in a multi-objective setting.
		How this utility function is defined has a significant impact on a multi-objective reinforcement learning problem should be approached.

		Unfortunately, how utility is derived from a satellite-based active fire detection system is a complicated matter touching on many areas of expertise.
		Factors that can be considered include:
		\begin{enumerate}
			\item The cost of false positives compared to the cost of false negatives.
			This requires assessing the impact of dispatching firefighting resources to a location with no fire, versus not dispatching fire fighting resources to a location that has fire.
			\item The cost is of a certain area being burned.
			Wildfires are costly because of the damage they cause, but creating numerical models for the damage caused by a wildfire is complex and requires expertise in many areas.
			\item How quickly wildfires can be detected.
			Finding wildfires quickly is essential to controlling them, since it is much easier to control and extinguish a small wildfire compared to a larger fire.
		\end{enumerate}

		Building a utility function to compute the utility derived from the use of a particular neural network for the detection of fires requires expertise across a number of areas and is beyond the area of expertise of the authors.
		Given the primary goal of this paper of providing guidance on the use of a reinforcement learning-based \ac{NAS} search strategy to solve novel problems, we consider such in-depth analysis outside the scope of this publication.
		With this in mind, we opt to use a linear combination that values task-performance, and the required computational resources equally.
		The use of a linear combination as a utility function also significantly simplifies the multi-objective reinforcement learning problem considered.
		For example, under the assumption of a linear utility function the two different optimization criteria outlined by \textcite{Radulescu_2019_Multi} (\ac{SER} and \ac{ESR}) become equivalent.

		The specific reinforcement learning method we use is the method first described by \textcite{Cassimon_2024_Scalable}.
		They evaluate a transformer-based reinforcement learning agent trained using Ape-X~\cite{Horgan_2018_Distributed}, a variant of deep Q-learning, on two \ac{NAS} benchmarks: NAS-Bench-101 and NAS-Bench-301.
		We use the same agent architecture with similar hyperparameters, but use a different reward function based on the linear combination of the post-quantization F1 score of the designed architecture and the number of trainable parameters in the designed neural network.

\section{Experiments}
\label{sec:experiments}
	In this Section, we outline the data collected and the different experiments performed in the context of this paper.
	We start with a brief analysis of the performance of different neural networks on the target task in Section~\ref{subsec:experiments:perf_data_anal}.
	Next, we analyze the process of training a supervised performance prediction model on the collected performance data in Section~\ref{subsec:experiments:perf_pred_training}.
	We follow this with an analysis of the performance of our reinforcement learning agent on the \ac{NAS} problem in Section~\ref{subsec:experiments:nas_agent}.
	Finally, in Section~\ref{subsec:experiments:deployment}, we select the best model found by our reinforcement learning agent and deploy it on a Google Coral Micro Dev Board, we conduct some measurements regarding inference performance and power consumption to assess the viability of using our neural network-based approach in a nanosatellite context.

	\subsection{Statistical Analysis of Performance Data}
	\label{subsec:experiments:perf_data_anal}
		When gathering neural network performance data, up to 6 neural networks were trained concurrently using the same \ac{GPU}.
		The neural networks were trained using the \ac{SGD} optimizer~\cite{Rosenblatt_1958_Perceptron} with a learning rate of $1 \times 10^{-2}$ and Nesterov momentum with a weight of $0.9$.
		Gradients norms were clipped to 20, and we applied kurtosis regularization~\cite{Shkolnik_2020_Robust} to aid in quantization with a weight of $1 \times 10^{-2}$ and a target of kurtosis of 1.8.
		The neural networks also had L2 regularization applied with a weight of $1 \times 10^{-2}$.
		The networks were trained with a batch size of $16,384$ for $500$ epochs.

\begin{figure}
			\centering
			\includegraphics[width=\columnwidth, clip]{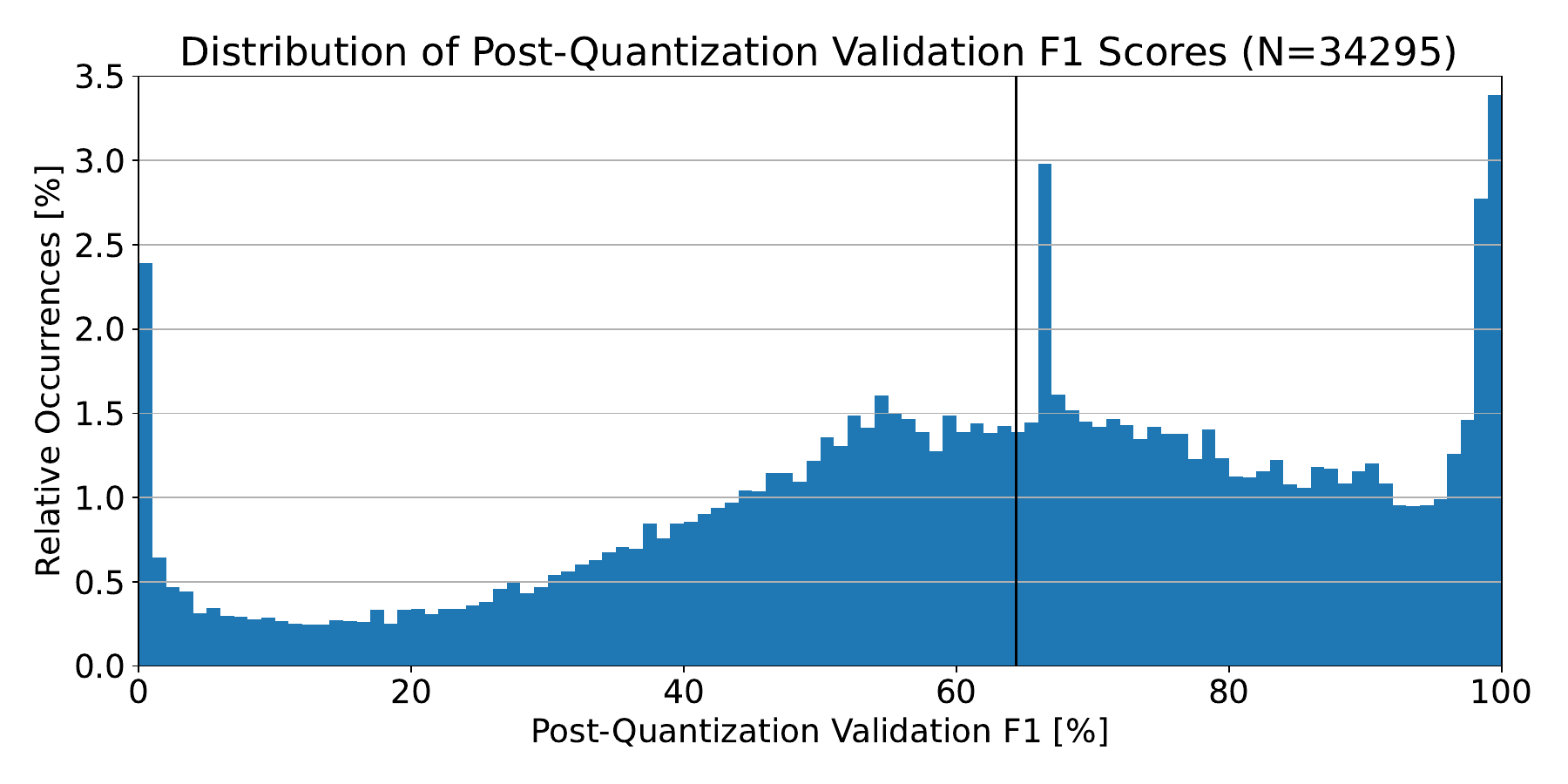}
			\caption{A histogram showing the obtained post-quantization F1 scores on the validation set. The median is indicated with a solid black vertical line.}
			\label{fig:experiments:f1_histogram}
		\end{figure}

		First, we analyze the performance we attained on the task of detecting active fires from single multispectral images.
		As mentioned in Section~\ref{subsec:methods:performance_data_gathering}, we were able to train $34,295$ neural networks within our computational budget.
		In total, this dataset comprises of $11,547$ unique neural network architectures, each trained using three different random initializations, with seeds ranging from 0 to 2 (both inclusive).
		Version of TensorFlow Lite 2.15 does not always succeed in quantizing a neural network following training.
		Figure~\ref{fig:experiments:f1_histogram} shows a distribution of the post-quantization F1 score achieved on the validation set for all networks that we were able to successfully quantize.
		A solid black line indicates the median F1 score of 64.36\%.
		We note that a large percentage of architectures ended up in the first and last buckets (Buckets have a width of 1\%), indicating many architectures either have an accuracy in the $\left[0\%, 1\%\right[$ or the $\left[99\%, 100\%\right]$ range.

		In their paper, \textcite{Cassimon_2024_Scalable} note the necessity of reward shaping when rewards are concentrated in a small region of their possible range.
		They use an exponential reward shaping function to correct for the fact that the majority of the architectures in the NAS-Bench-101 dataset have accuracies around 90\%.
		In our use-case, reward shaping will not be necessary based on the data in Figure~\ref{fig:experiments:f1_histogram}.

		Figure~\ref{fig:experiments:f1_histogram} also shows that F1 scores in our use-case are not uniformly distributed.
		Given that bins in this histogram have a width of 1\%, we'd also expect the bars to have a height of 1\% under the assumption of a uniform distribution.
		This is not the case, with architectures with an F1-score below roughly 40\% being underrepresented, while architectures with an accuracy above 40\% being overrepresented.


	\subsection{Performance Predictor}
	\label{subsec:experiments:perf_pred_training}
		After gathering the performance data by training a large set of neural networks, we are ready to train our performance prediction models.
		As mentioned in Section~\ref{subsec:methods:performance_predictor_training}, we consider a wide variety of predictors.
		Table~\ref{tab:experiments:perf_pred_hparams} lists the hyperparameters for the  performance predictors.
		Both random samplers have no hyperparameters.
		Parameters that aren't specified use the default values in \verb|scikit-learn| 1.4.0~\cite{Pedregosa_2011_Scikit}.
		All models that accept a seed were given the same seed.
		The seed was a randomly selected integer in $\left[0, 2^{32}-1\right]$ selected from a \verb|numpy|~\cite{Harris_2020_Array} random number generator initialized with seed 0.
		Models that accept a ``n\_jobs'' argument were given $-1$ for ``n\_jobs''
		All linear models were configured to fit scale and intercept.

		\begin{table*}
			\centering
			\caption{Overview of the hyperparameters used to train the different performance predictors. \ac{OLS} was ommitted because we used the default hyperparameters.}
			\begin{tabular}{l | r ? l | r ? l | r}
				\toprule
				\multicolumn{2}{c?}{\textbf{Ridge Regression}}			& \multicolumn{2}{c?}{\textbf{\ac{LASSO}}}													& 							&\\
				\midrule
				$\alpha$								&	$1$					& $\alpha$								& $1 \times 10^{-3}$											& 							&\\
				\toprule
				\multicolumn{2}{c?}{\textbf{Multi-Layer Perceptron}}	& \multicolumn{2}{c?}{\makecell{\textbf{Radius Nearest}\\\textbf{Neighbour Regression}}}	& \multicolumn{2}{c}{\makecell{\textbf{K Nearest}\\\textbf{Neighbour Regression}}}\\
				\midrule
				$\mathit{hidden\_layer\_sizes}$	& $\left[48, 48\right]$	& $\mathit{radius}$					& $16$													& $\mathit{n\_neighbours}$	& $100$\\
				$\alpha$								& $1 \times 10^{-3}$			& $\mathit{weights}$				& $\mathit{distance}$									& $\mathit{weights}$		& $\mathit{uniform}$\\
				\toprule
				\multicolumn{2}{c?}{\textbf{Gradient Boosted Trees}}	& \multicolumn{2}{c?}{\textbf{\ac{SGD}}}													& \multicolumn{2}{c}{\textbf{Gaussian Processes}}\\
				\midrule
				$\mathit{n\_estimators}$			& $75$					& $\mathit{max\_iter}$				&	$5000$												& $\mathit{kernel}$			& $32 \times \exp\left(-\frac{d\left(x_{i},  x_{j}\right)^{2}}{2 \cdot 16^{2}}\right)$\\
				$\mathit{max\_depth}$			& $4$					& $\eta_{0}$							&	$1 \times 10^{-5}$										& $\alpha$					& $1 + \sigma^{2}$\\
				$\mathit{max\_leaves}$			& $8$					& $\mathit{learning\_rate}$			&	$\mathit{constant}$									& $\mathit{normalize\_y}$	& $\mathit{True}$\\
				\toprule
				\multicolumn{2}{c?}{\textbf{\ac{GNN}}}					& \multicolumn{2}{c?}{\textbf{Random Forest}}												& \multicolumn{2}{c}{\textbf{\ac{SVM}}}\\
				\midrule
				$\mathit{hidden\_layer\_sizes}$	& $\left[64, 64\right]$	& $\mathit{n\_estimators}$			&	$100$												& $\mathit{kernel}$			& $\mathit{rbf}$\\
				$\mathit{activation}$			& $\mathit{Leaky ReLU}$	& $\mathit{max\_depth}$				&	$15$												& $C$						& $5$\\
				$\eta_{0}$							& $1 \times 10^{-2}$			& $\mathit{min\_samples\_split}$		&	$50$												& $\gamma$						& $\mathit{auto}$\\
				$\eta_{\gamma}$							& $2.5 \times 10^{-1}$		& $\mathit{min\_samples\_leaf}$		&	$25$												& $\epsilon$						& $1 \times 10^{-2}$\\
				\bottomrule
			\end{tabular}
			\label{tab:experiments:perf_pred_hparams}
		\end{table*}

		Our \ac{GNN} is based on the graph convolutions first introduced by \textcite{Kipf_2017_Semisupervised}.
		The \ac{DAG} of each architecture is given as input and fed through two \ac{GCN} layers with leaky ReLU~\cite{Maas_2013_Rectifier} activations after each.
		After the final \ac{GCN} layer, the features of each node are accumulated into one vector for the entire graph.
		The feature vector for the graph is finally fed through a linear layer resulting in a single scalar output.
		The \ac{GNN} was trained using the Adam optimizer~\cite{Kingma_2014_Adam} minimizing the \ac{RMSE} over 100 epochs.
		The learning rate started out at $\eta_{0}=0.01$, and was multiplied by $\eta_{\gamma}=0.25$ after the 5th, 10th and 50th epoch.

		\begin{figure}
			\centering
			\includegraphics[width=\columnwidth, clip]{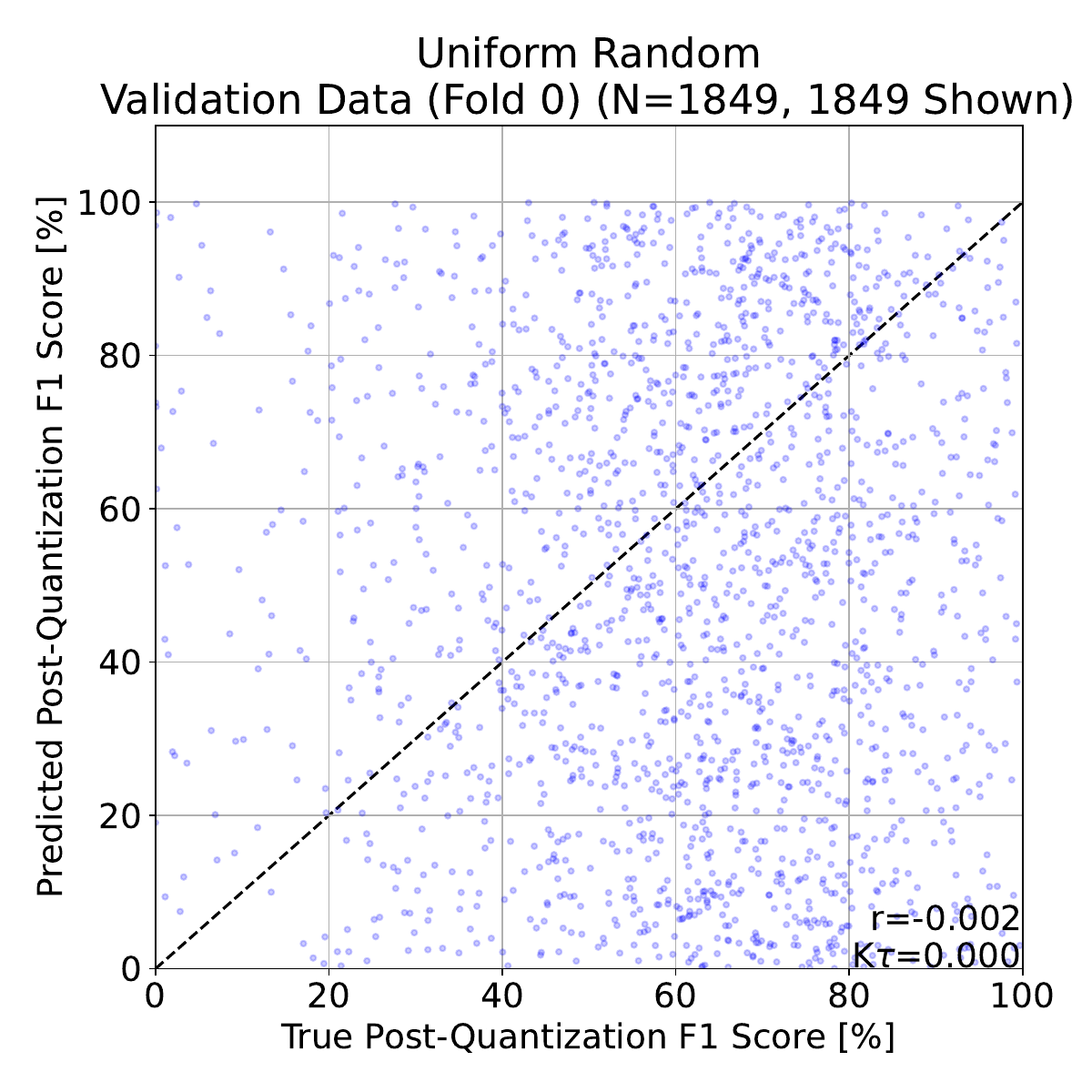}
			\caption{Scatter plot for predictions from the uniform model.}
			\label{fig:experiments:uniform-random:validation}
		\end{figure}

		\begin{figure}
			\centering
			\includegraphics[width=\columnwidth, clip]{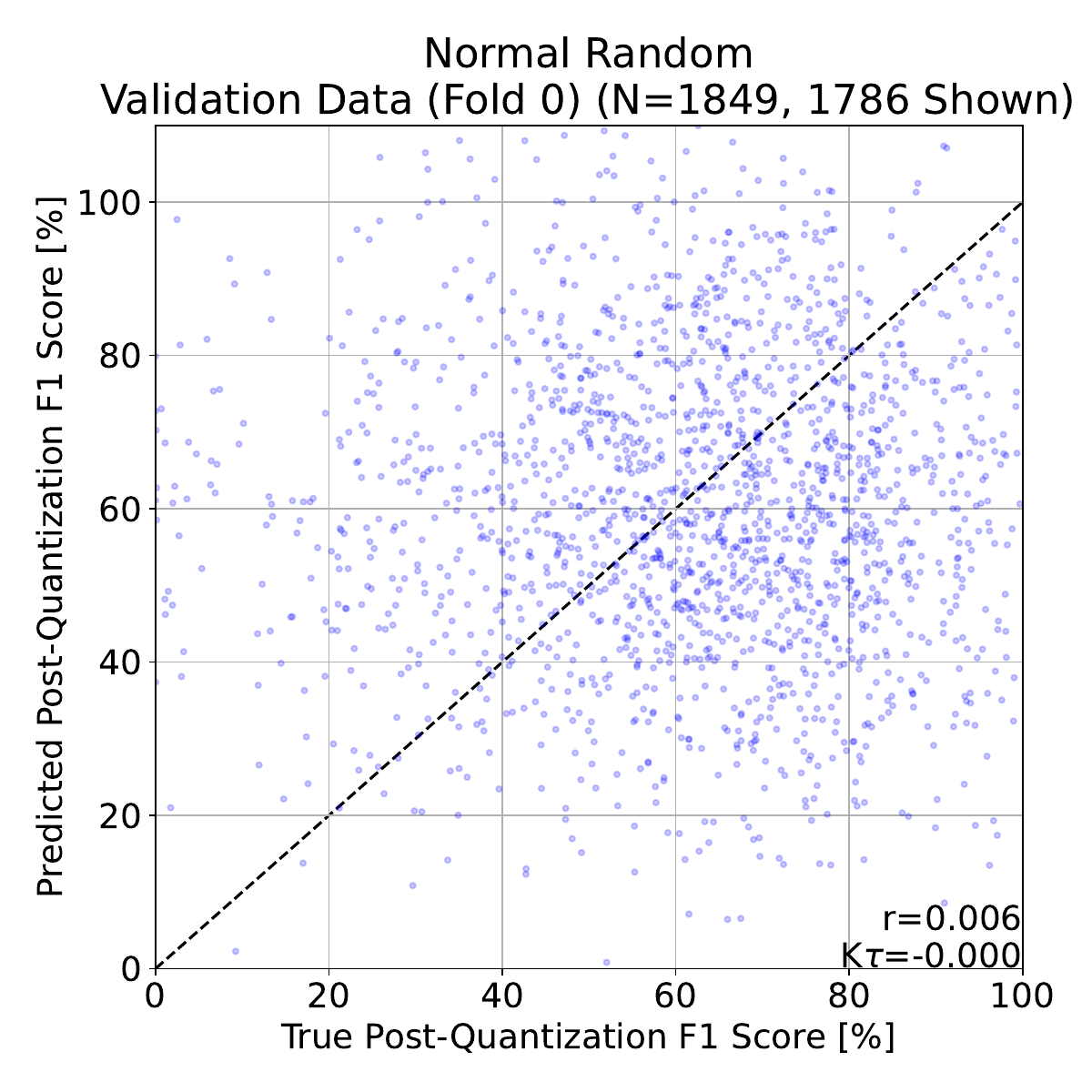}
			\caption{Scatter plot for predictions from the normal model.}
			\label{fig:experiments:normal-random:validation}
		\end{figure}

%
%

		Table~\ref{tab:experiments:perf-pred-results} shows the numerical results for performance predictors, sorted by \ac{RMSE} on the validation set.

		First, we consider predictions on the validation set by both random predictors, to provide a frame of reference when analyzing other predictors.
		Figure~\ref{fig:experiments:uniform-random:validation} shows the scatter plot for a uniform random sampler.
		There is little remarkable about this figure, but one interesting thing to note is that the density of the scatter plot nicely reflects the distribution seen in Figure~\ref{fig:experiments:f1_histogram}.
		Figure~\ref{fig:experiments:normal-random:validation} shows the same scatter plot, but from our model based on a normal distribution.
		We note that in this case, the model generated several predictions outside the range shown on the graph.
		This is reflected in the title of the graph: Of the $1,849$ datapoints, only $1,786$ are within the range shown on the graph.

		\begin{figure}
			\centering
			\includegraphics[width=\columnwidth, clip]{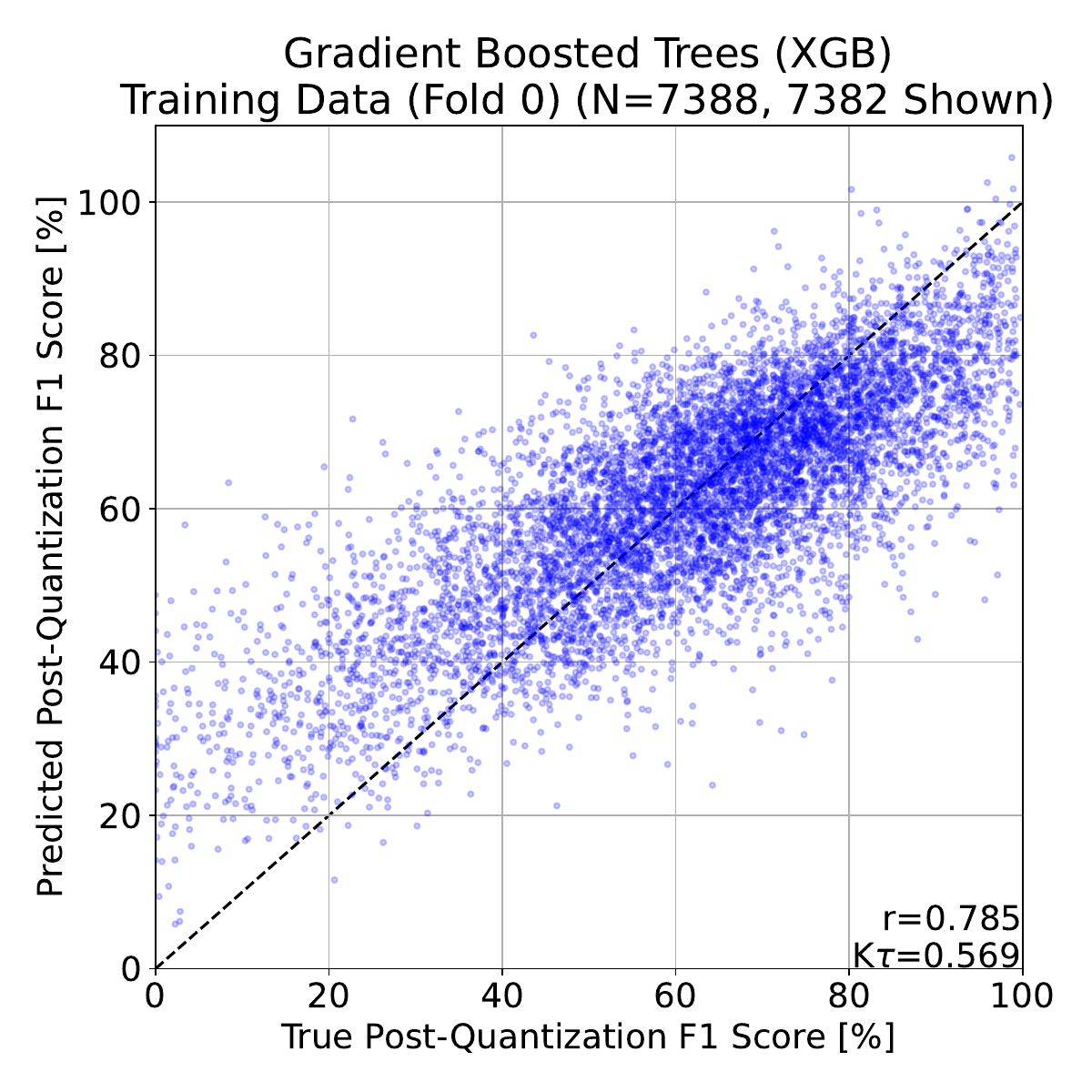}
			\caption{Scatter plot for predictions from the gradient boosted trees model - Training set.}
			\label{fig:experiments:xgb-predictions:training}
		\end{figure}

		\begin{figure}
			\centering
			\includegraphics[width=\columnwidth, clip]{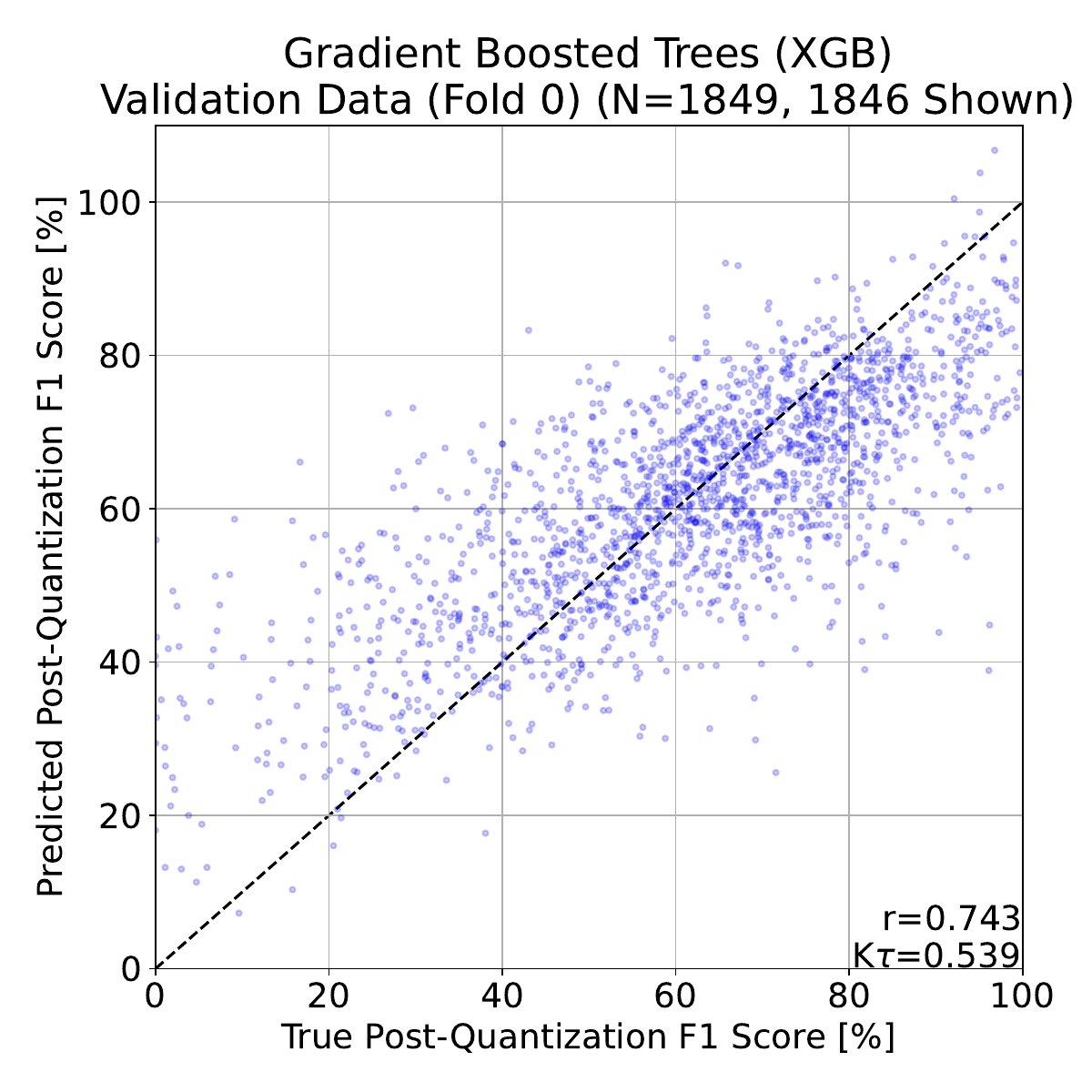}
			\caption{Scatter plot for predictions from the gradient boosted trees model - Validation set.}
			\label{fig:experiments:xgb-predictions:validation}
		\end{figure}


		Next, we consider the results for the best performing model, gradient boosted trees.
		We show a scatter plot for the first fold of both the training set (Figure~\ref{fig:experiments:xgb-predictions:training}) and the validation set (Figure~\ref{fig:experiments:xgb-predictions:validation}), and we note that other folds show similar results.
		The predictions capture the overall trend, but are far from precise.
		The predictions still display a significant error that is often between $-20\%$ and $+20\%$.
		This is also reflected in Table~\ref{tab:experiments:perf-pred-results}, which shows \iac{RMSE} on the validation set of 13.7\% for gradient boosted trees.
		Interesting to note on Figure~\ref{fig:experiments:xgb-predictions:training} in particular is the difference in prediction accuracy between samples with a ground truth F1 score below 40\% and samples with a ground truth F1 score above 40\%.
		This is likely a consequence of the observation we made in Section~\ref{subsec:experiments:perf_data_anal} about Figure~\ref{fig:experiments:f1_histogram} that architectures with an F1 score below 40\% are underrepresented in our dataset.

\begin{figure}
			\centering
			\includegraphics[width=\columnwidth, clip]{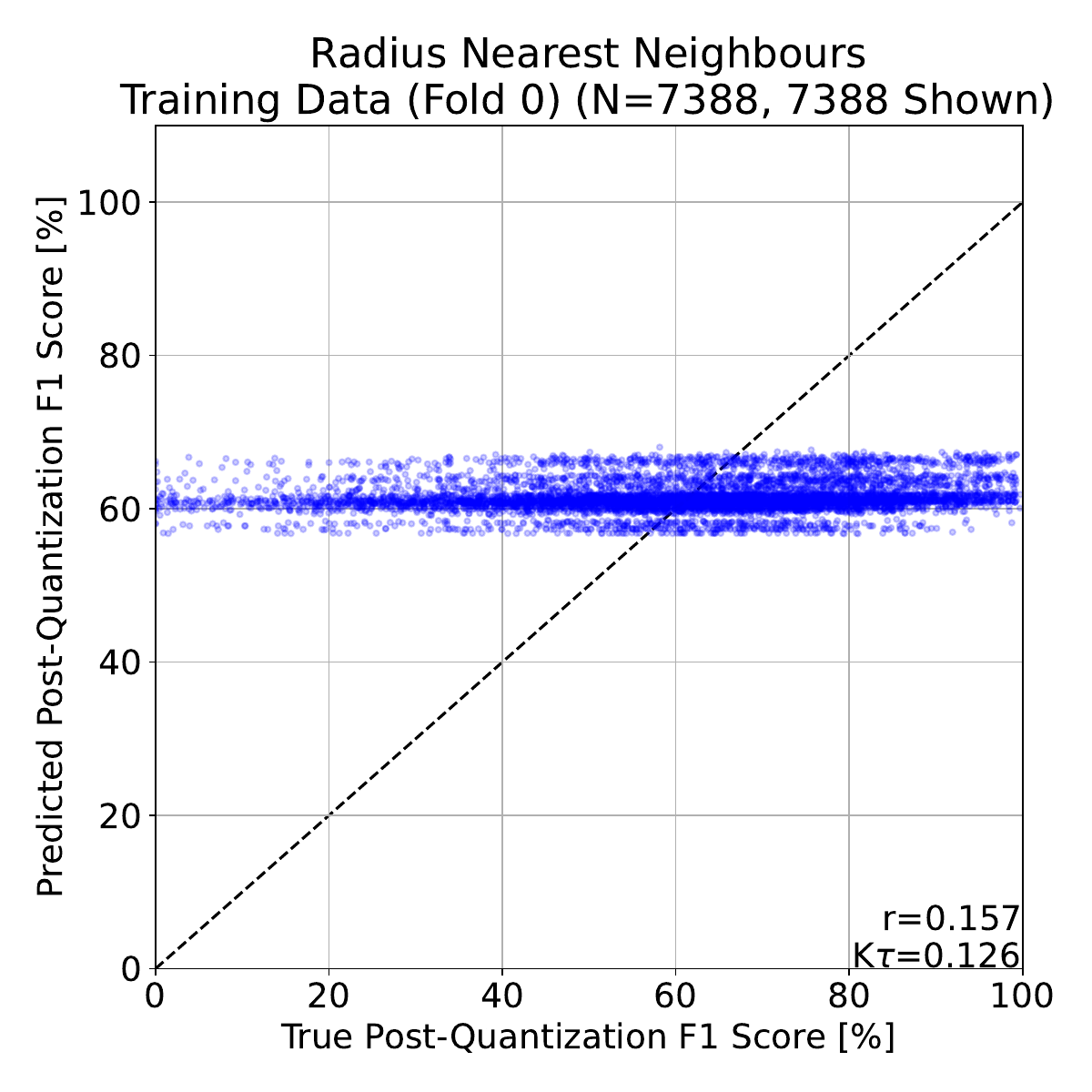}
			\caption{Scatter plot for predictions from the radius nearest neighbours model - Training set.}
			\label{fig:experiments:radius-predictions:training}
		\end{figure}

\begin{figure}
			\centering
			\includegraphics[width=\columnwidth, clip]{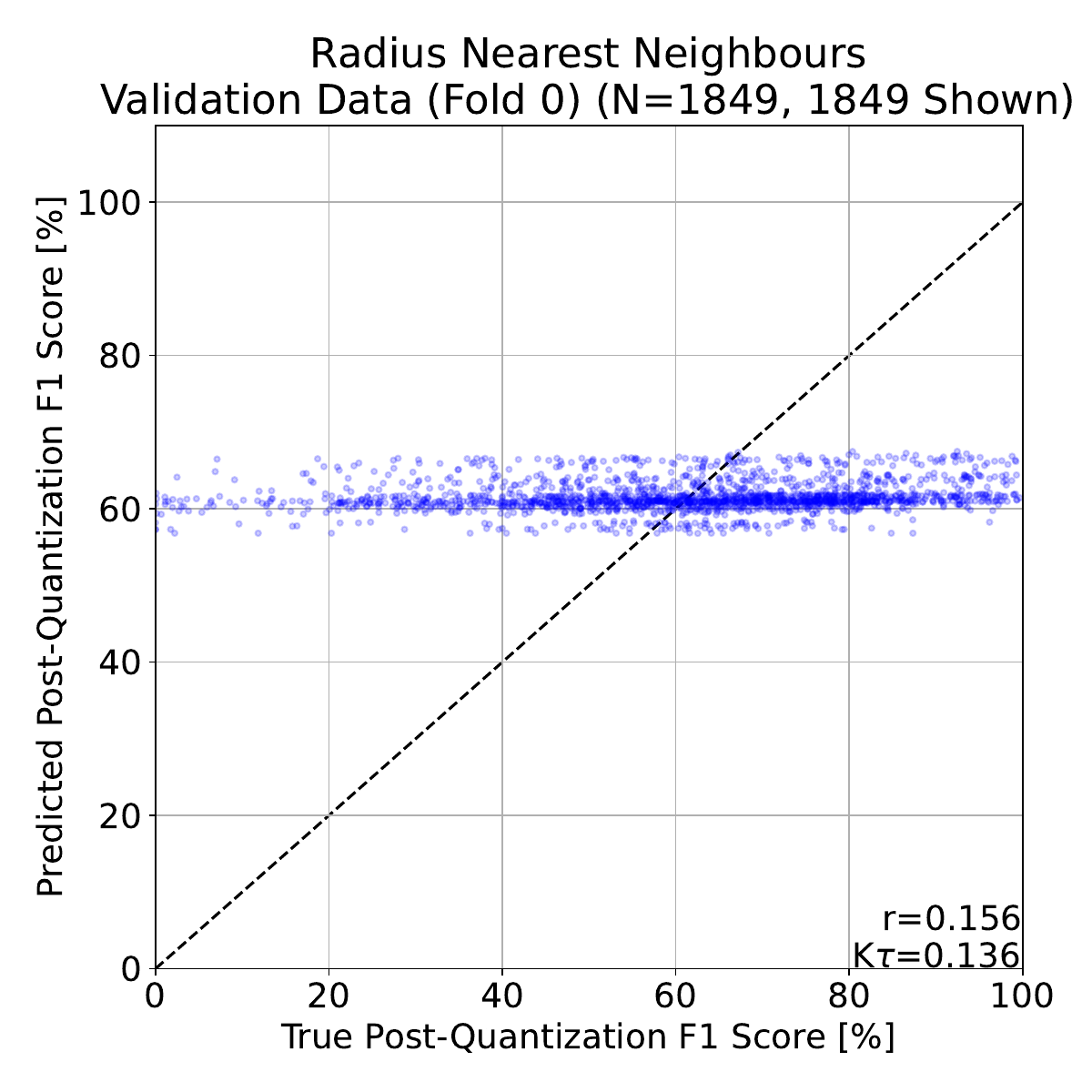}
			\caption{Scatter plot for predictions from the radius nearest neighbours model - Validation set.}
			\label{fig:experiments:radius-predictions:validation}
		\end{figure}

		From Table~\ref{tab:experiments:perf-pred-results} we see that most predictors managed to produce a reasonably good model.
		Excluding the random models, the radius nearest neighbour predictor is the only model that has a Pearson's R below 50\%.
		We considered several distance metrics, radii and tested both uniform and distance weights, but were unable to find a configuration that yielded a good fit.
		When using distance weights, the model overfit very easily, while when using uniform weights, the model usually underfit.
		Figure~\ref{fig:experiments:radius-predictions:training} and~\ref{fig:experiments:radius-predictions:validation} show predictions made by the radius nearest neighbour model on the training and validation dataset.
		Using this configuration, it is clear that the model underfit the training dataset.
		Instead, the predictions seem to vary (narrowly) around the mean post-quantization F1 score of $64\%$ we show in Figure~\ref{fig:experiments:f1_histogram}.

		\begin{table*}
			\centering
			\caption{A Comparison of the different performance prediction models. Each column is given as a mean and standard deviation over 5 folds. The algorithm with the best mean performance in each column has been marked in bold. Numbers marked with an asterisk excluded some outliers caused by poor convergence of the algorithm. All statistics were computed using the validation set, unless stated otherwise.}
			\begin{tabular}{l | r | r | r | r| r}
				\toprule
				Algorithm							& Pearson's R							& Kendall's $\tau$						& $R^{2}$								& \ac{RMSE}								& \ac{RMSE} (Training Set)	\\
				\midrule
				Gradient Boosted Trees			   & $\mathbf{74.2\% \pm  0.58\%}$			& $\mathbf{53.1\% \pm  0.80\%}$			& $\mathbf{54.9\% \pm  0.74\%}$			& $\mathbf{13.7\% \pm  0.17\%}$			& $12.7\% \pm  0.02\%$					  \\
				Support Vector Machine			   & $73.9\% \pm  0.41\%$					 & $53.1\% \pm  0.54\%$					 & $54.4\% \pm  0.68\%$					 & $13.8\% \pm  0.18\%$					 & $\mathbf{11.3\% \pm  0.07\%}$			 \\
				Gaussian Process					 & $73.2\% \pm  0.28\%$					 & $52.3\% \pm  0.69\%$					 & $53.5\% \pm  0.40\%$					 & $13.9\% \pm  0.13\%$					 & $13.3\% \pm  0.03\%$					  \\
				Ridge Regression					 & $70.4\% \pm  0.28\%$					 & $49.9\% \pm  0.70\%$					 & $49.5\% \pm  0.44\%$					 & $14.5\% \pm  0.15\%$					 & $14.4\% \pm  0.04\%$					  \\
				Ordinary Least Squares			   & $56.8\% \pm 27.25\%$					 & $49.9\% \pm  0.70\%$					 & $49.5\% \pm  0.47\%^{*}$				 & $14.6\% \pm  0.08\%^{*}$				 & $14.4\% \pm  0.04\%$					  \\
				SGD Regression					   & $70.0\% \pm  0.23\%$					 & $49.5\% \pm  0.61\%$					 & $48.5\% \pm  0.26\%$					 & $14.7\% \pm  0.16\%$					 & $14.5\% \pm  0.03\%$					  \\
				LASSO Regression					 & $69.7\% \pm  0.37\%$					 & $49.2\% \pm  0.76\%$					 & $47.9\% \pm  0.44\%$					 & $14.7\% \pm  0.12\%$					 & $14.6\% \pm  0.04\%$					  \\
				Multi-Layer Perceptron			   & $70.7\% \pm  1.22\%$					 & $50.0\% \pm  1.20\%$					 & $47.6\% \pm  2.06\%$					 & $14.8\% \pm  0.27\%$					 & $11.4\% \pm  0.37\%$					  \\
				Random Forest						& $67.5\% \pm  0.80\%$					 & $47.5\% \pm  0.76\%$					 & $45.4\% \pm  1.03\%$					 & $15.1\% \pm  0.11\%$					 & $14.4\% \pm  0.02\%$					  \\
				K Nearest Neighbours				 & $61.3\% \pm  1.15\%$					 & $42.0\% \pm  0.77\%$					 & $28.3\% \pm  0.72\%$					 & $17.3\% \pm  0.20\%$					 & $17.0\% \pm  0.04\%$					  \\
				Graph Neural Network				 & $51.9\% \pm  2.34\%$					 & $35.6\% \pm  1.60\%$					 & $23.7\% \pm  1.39\%$					 & $17.8\% \pm  0.30\%$					 & $17.8\% \pm  0.18\%$					  \\
				Radius Nearest Neighbours			& $15.4\% \pm  1.81\%$					 & $12.6\% \pm  1.66\%$					 & $ 2.0\% \pm  0.32\%$					 & $20.2\% \pm  0.22\%$					 & $20.2\% \pm  0.05\%$					  \\
				Normal Random						& $ 0.0\% \pm  1.94\%$					 & $-0.5\% \pm  1.43\%$					 & $-102.6\% \pm  5.17\%$				   & $29.1\% \pm  0.30\%$					 & $29.0\% \pm  0.15\%$					  \\
				Uniform Random					   & $ 0.7\% \pm  2.37\%$					 & $ 0.4\% \pm  1.38\%$					 & $-237.1\% \pm  6.35\%$				   & $37.5\% \pm  0.46\%$					 & $37.0\% \pm  0.12\%$					  \\
				\bottomrule
			\end{tabular}
			\label{tab:experiments:perf-pred-results}
		\end{table*}

\begin{figure}
			\centering
			\includegraphics[width=\columnwidth, clip]{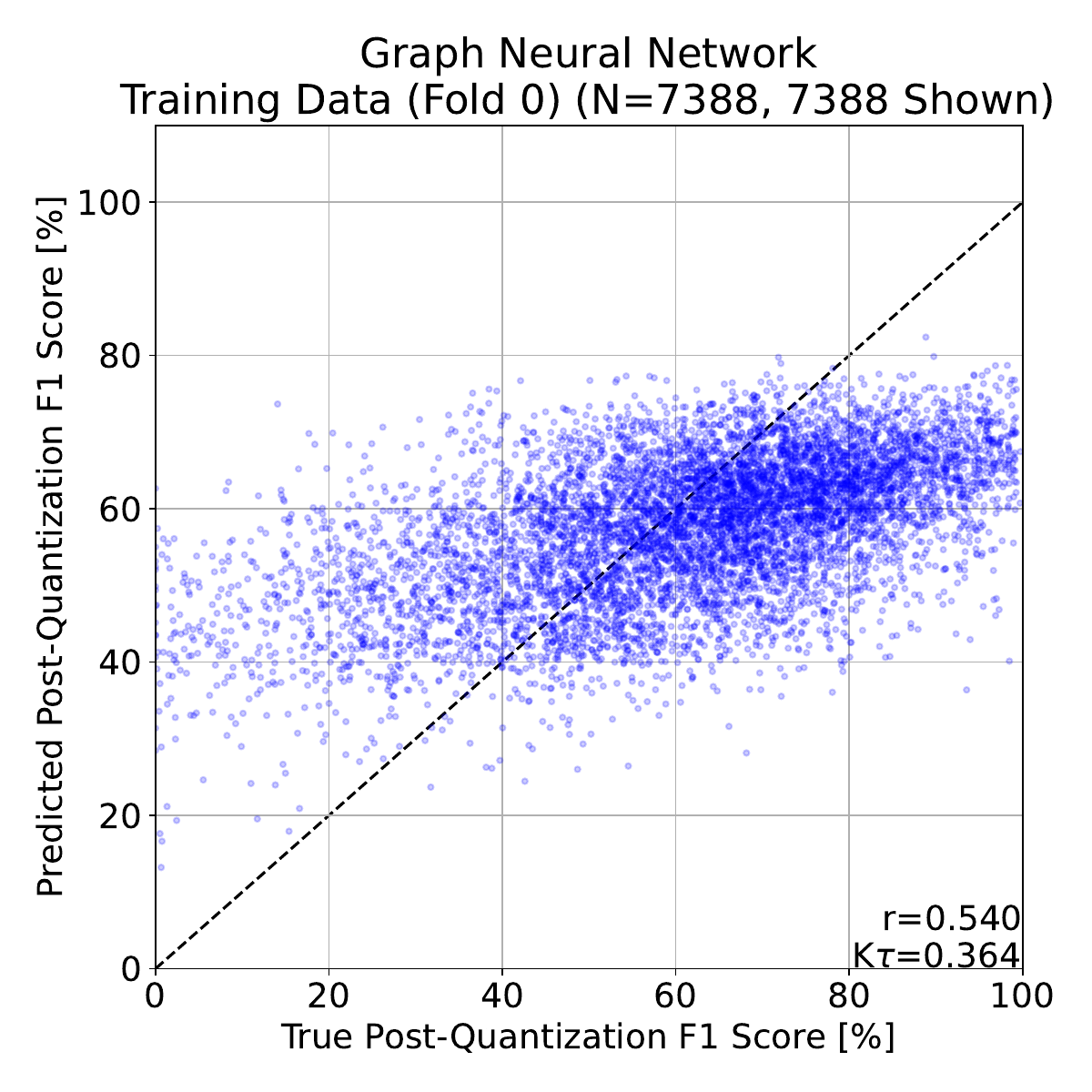}
			\caption{Scatter plot for predictions from the radius nearest neighbours model - Training set.}
			\label{fig:experiments:gnn-predictions:training}
		\end{figure}

\begin{figure}
			\centering
			\includegraphics[width=\columnwidth, clip]{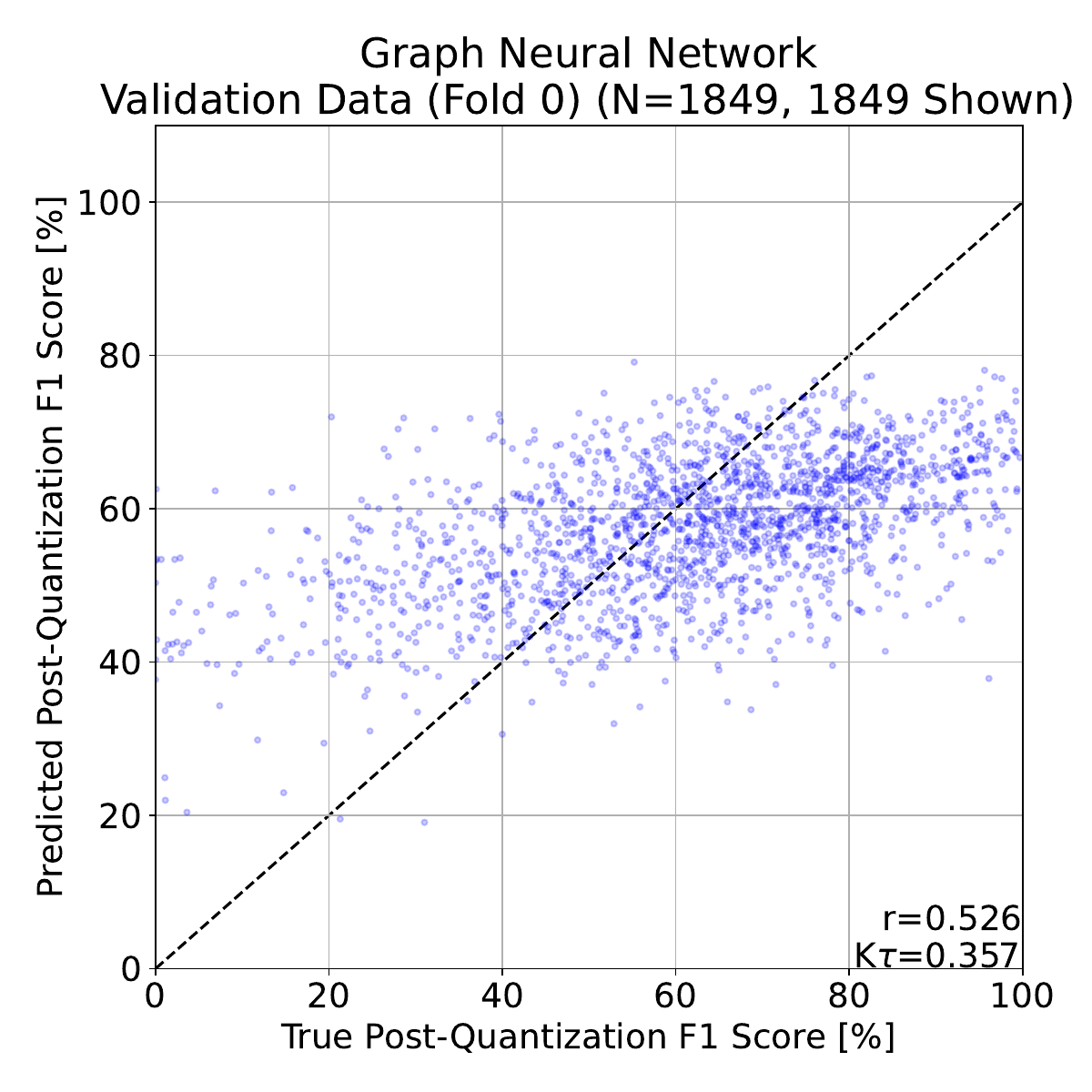}
			\caption{Scatter plot for predictions from the radius nearest neighbours model - Validation set.}
			\label{fig:experiments:gnn-predictions:validation}
		\end{figure}

		Finally, we also note the behaviour of the \ac{GNN}.
		Despite being intrinsically suited for dealing with graph data, the \ac{GNN} showed relatively weak performance.
		The \ac{GNN} seems to exhibit a much more extreme version of the bias displayed by the gradient boosted trees in Figures~\ref{fig:experiments:xgb-predictions:training}~and~\ref{fig:experiments:xgb-predictions:validation}.
		Figures~\ref{fig:experiments:gnn-predictions:training}~and~\ref{fig:experiments:gnn-predictions:validation} shows the predictions for the \ac{GNN} model on the training (Figure~\ref{fig:experiments:gnn-predictions:training}) and validation (Figure~\ref{fig:experiments:gnn-predictions:validation}) set.
		The model rarely predicts F1 scores below 40\%, reflecting the bias in the underlying distribution shown in Figure~\ref{fig:experiments:f1_histogram}.

	\subsection{NAS Agent}
	\label{subsec:experiments:nas_agent}
		In this section, we cover the details of the reinforcement learning agent that was used to design the final classification network for active fire detection.
		Following the experiments in Section~\ref{subsec:experiments:perf_pred_training}, we selected the gradient boosted tree regressor to be used as part of the reward function for our reinforcement learning, given its strong performance in predicting the post-quantization F1 score of the trained neural networks.
		As mentioned in Section~\ref{subsec:methods:nas_agent_training}, the reward function we use is a combination of the post-quantization F1 score predicted by the gradient boosted tree model and the total number of trainable parameters in the designed neural network.
		The total number of trainable parameters is first normalized between 0 and 1 before the linear combination is applied, to ensure both rewards have a similar magnitude.
		While it is relatively straightforward to define a lower bound for the total number of trainable parameters, finding an upper bound is harder.
		Since both bounds are necessary to compute the normalized total parameter count, we make an assumption on the worst-case.
		We assume that an architecture with the maximum number of edges and vertices, where every node is assigned the ``linear-prelu'' label is the worst-case in terms of total parameter count.
		This assumption is justified by the fact that for the ``linear-prelu'' operation not only the linear layer but also the activation function has trainable parameters.
		Finally, both components of the reward function are assigned an equal weight of $0.5$, and combined using a linear combination.

		We use the reinforcement learning agent introduced in \textcite{Cassimon_2024_Scalable} to design neural networks in an iterative fashion.
		The agent is given an architecture, along with a set of architectures obtained by modifying the first architecture as input.
		The agent then outputs the index of the preferred architecture.
		If the agent selects the current architecture, the episode terminates.
		If the agent selects a modified architecture, this architecture becomes the new current architecture and the process repeats.
		During training, episodes have a length of up to 16 time steps, with episodes taking up to 32 time steps during evaluation.
		Each agent is trained 5 times, with numbers from 0 to 4 (both inclusive) being used as seeds for random number generation.
		Training is terminated once the agent has been trained on $10 \times 10^{6}$ time steps of experience.
		We use $\gamma=0.9$ following the ablation study in \textcite{Cassimon_2024_Scalable}.
		Agents are trained using the Ape-X algorithm~\cite{Horgan_2018_Distributed}, a variant of deep Q learning.
		Agents are shown up to $50$ neighbours each time step.
		We use the Adam optimizer~\cite{Kingma_2014_Adam} with a learning rate of $5 \times 10^{-5}$.
		We use double Q-learning and duelling heads with target networks that are updated every time the online network is trained for $8,192$ samples.
		Exploration is handled using a per-worker epsilon-greedy strategy: Every worker uses an epsilon-greedy exploration strategy with a different value of epsilon.
		For the precise calculation of epsilon values we refer to the original Ape-X paper~\cite{Horgan_2018_Distributed}.
		Following \textcite{Cassimon_2024_Scalable}, we use a replay buffer with $2.5 \times 10^{4}$ entries, using prioritized replay with $\alpha=0.6$ and $\beta=0.4$.
		Following the results from Section~\ref{subsec:experiments:perf_data_anal} and contrary to \textcite{Cassimon_2024_Scalable}, we do not use any reward shaping.
		When evaluating, we randomly sample 5 sets of $1 \times 10^{4}$ architectures from the search space and evaluate the agents on each of these sets.
		We also include random search, random walks and local search as additional baselines.

		\begin{figure}
			\centering
			\includegraphics[width=\columnwidth, clip]{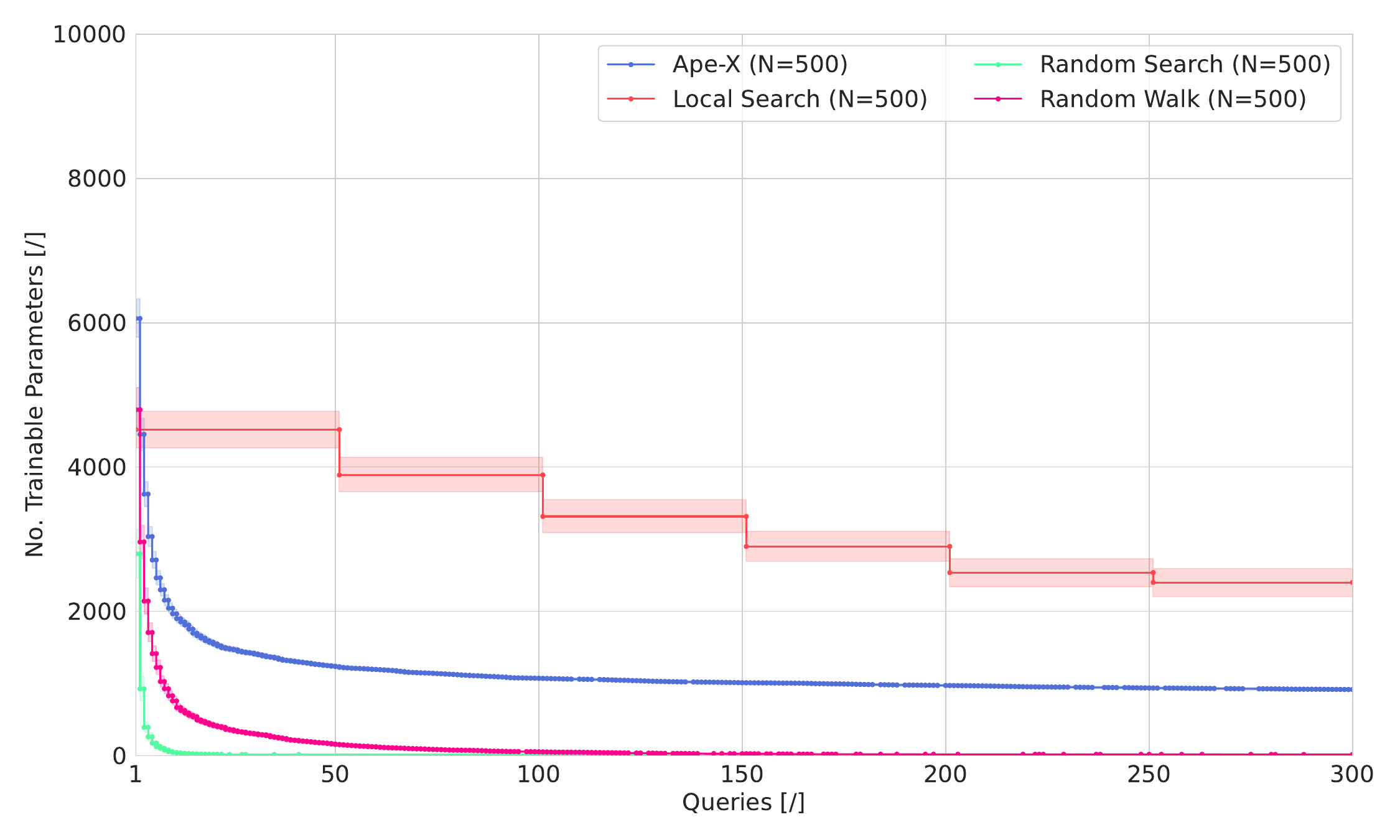}
			\caption{The lowest number of trainable parameters obtained as a function of the number of queries made.}
			\label{fig:nas-results:best-tpc-after-N}
		\end{figure}

\begin{figure}
			\centering
			\includegraphics[width=\columnwidth, clip]{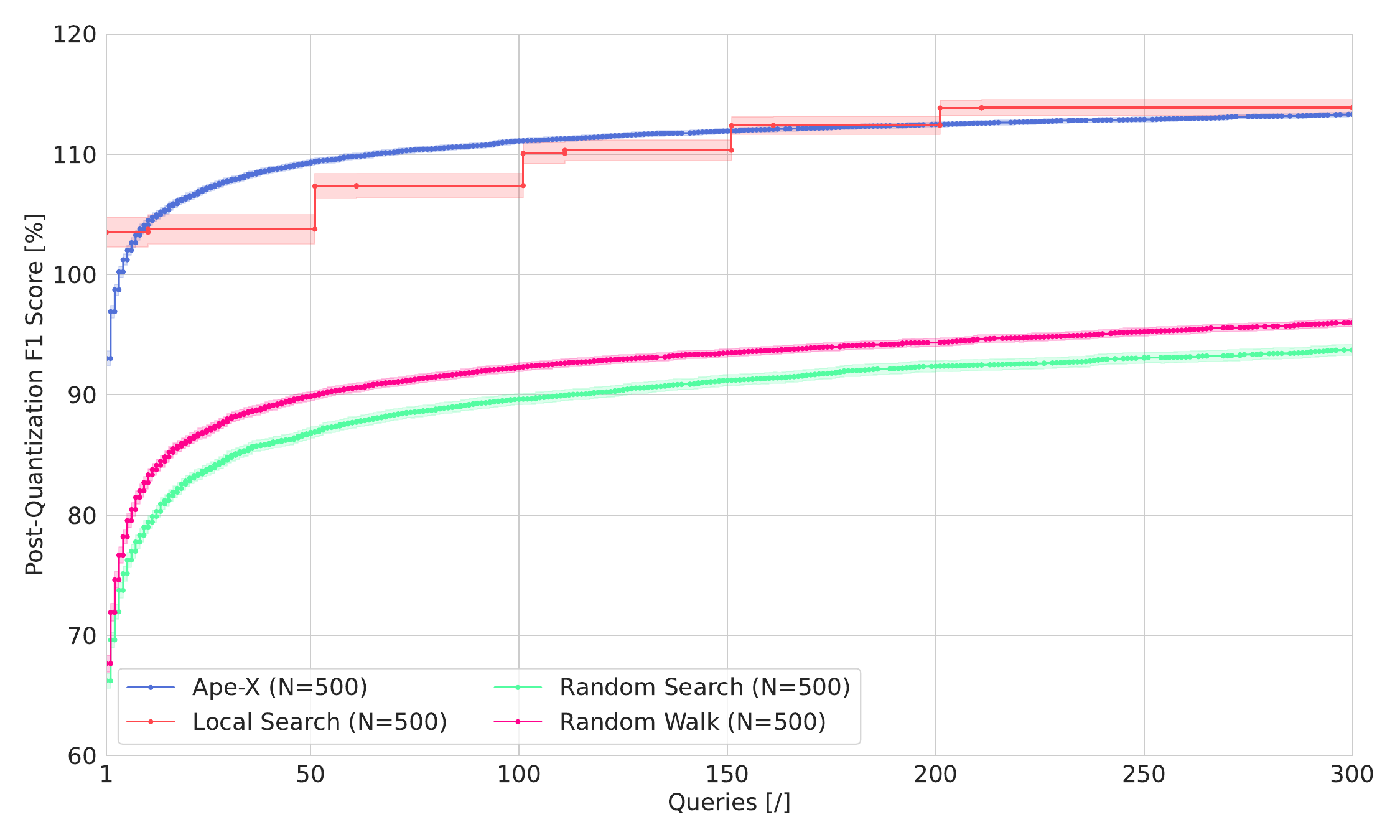}
			\caption{The highest post-quantization F1-score obtained as a function of the number of queries made.}
			\label{fig:nas-results:best-f1-after-N}
		\end{figure}

		Figures~\ref{fig:nas-results:best-f1-after-N} and~\ref{fig:nas-results:best-tpc-after-N} show the performance of all agents on both objectives as a function of the number of queries.
		One thing that immediately stands out in Figure~\ref{fig:nas-results:best-tpc-after-N} is how easy it is for the random search agent to design neural networks with very low parameter counts.
		While this may seem surprising at first, it follows from the sampling algorithm used for random search.
		We use the same sampling strategy detailed in \textcite{Cassimon_2024_Scalable}.
		As shown in Figure 5 in \textcite{Cassimon_2024_Scalable}, the random search algorithm samples uniformly in function of vertex counts.
		Since we consider architectures with 2 to 8 vertices (both inclusive), random search has a roughly 1/6 chance ($16.66\ldots\%$) to sample an architecture with only 2 vertices.
		An architecture with only 2 vertices has exactly one edge and no trainable parameters, apart from the linear projection used in the output.
		Thus, such an architecture has 17 trainable parameters, and occurs with a $16.66\ldots\%$ probability at the end of an episode.
		This also explains why random search struggles much more to find architectures with high post-quantization F1 scores, given that these are much rarer.

		Figure~\ref{fig:nas-results:best-f1-after-N} also shows an interesting artifact.
		Both local search and our reinforcement learning agent are able to find architectures with post-quantization F1 scores above 100\% fairly easily.
		This indicates that both algorithms have managed uncover adversarial inputs to the performance prediction model, that lead it to predict an impossible F1 score.
		We saw a limited number of samples that were predicted to have F1 scores above 100\% in Figure~\ref{fig:experiments:xgb-predictions:validation}, but they are much more frequent in the evaluation data.
		Specifically, such adversarial samples are most common with local search, occurring in $246,755$ traces ($98.70\%$), followed by our reinforcement learning agent with $39,925$ ($15.97\%$) occurrences.
		They occur only rarely with either random algorithm, with random search reporting $95$ occurrences ($0.04\%$) and random walks reporting $136$ ($0.05\%$) occurrences.
		We hypothesize that these adversarial samples likely adversely impacted the performance of the reinforcement learning agent to a limited degree.
		A possible mitigation strategy is to simply clamp the predictions from the performance prediction model between 0 and 1 (Or 100\%).
		The impact of such a mitigation measure may be limited however, since it does not address the underlying cause of the adversarial samples (An imperfect regression model), rather it only address the symptom (Overprediction of post-quantization F1 scores).
		Building stronger performance prediction models is likely the best mitigation strategy against such issues.

		\begin{figure*}
			\centering
			\includegraphics[width=\textwidth, clip]{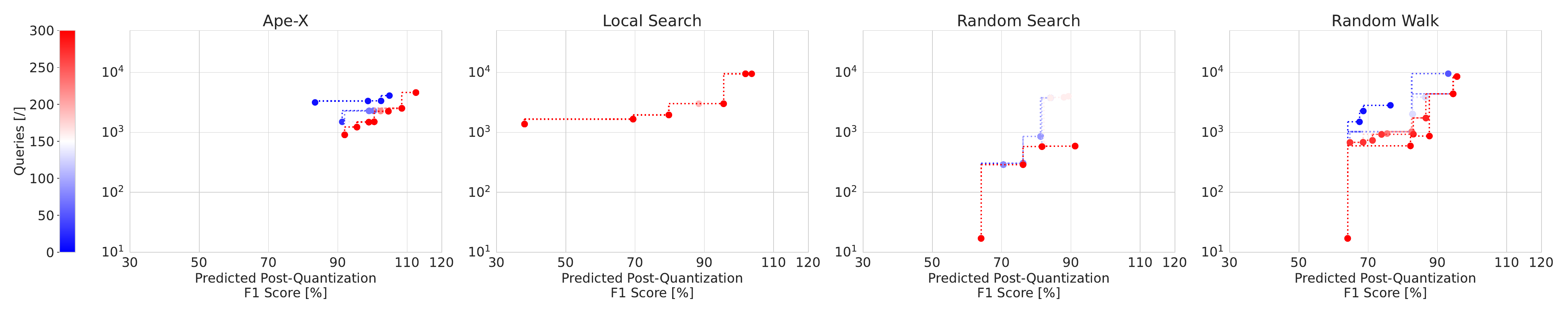}
			\caption{Pareto fronts generated by each agent given a query budget of 300 queries. Every 10 queries, it is redrawn. The first pareto front is drawn in blue, the last in red, and others interpolate between blue and red based on how many queries were required to find the front. Note the logarithmic Y axis.}
			\label{fig:experiments:pareto-fronts}
		\end{figure*}

		Figure~\ref{fig:experiments:pareto-fronts} shows the pareto fronts found by each agent under a query budget of 300 queries.
		These pareto fronts confirm what was already visible in Figures~\ref{fig:nas-results:best-f1-after-N} and~\ref{fig:nas-results:best-tpc-after-N}, that random algorithms don't struggle with finding neural networks with low parameter counts, but they do struggle with finding neural networks  with high post-quantization F1 scores.
		We count queries in the same way as \textcite{Cassimon_2024_Scalable}, which severely limits the number of opportunities local search has for making improvements to the architecture.
		In this search space, most architectures have 50 neighbours, allowing local search 6 time steps to make improvements before its query budget is exhausted.
		Comparatively, other algorithms can play up to 300 episodes (Up to 4800 time steps), since they only need to query an architecture's accuracy at the end of the episode.
		This likely also explains why local search did comparatively worse than Ape-X, with Ape-X finding better architectures both in terms of the total number of trainable parameters and the predicted post-quantization F1 score.
		Ape-X's pareto front is also relatively compact compared to those of the random algorithms, this is likely explained by the fact that random search (And local search) try to honour the preference vector we selected ($\left[0.5, 0.5\right]$), whereas random search and random walks simply sample randomly, without regard for the preference vector used.

	\subsection{Deployment}
	\label{subsec:experiments:deployment}
		\begin{figure}
			\centering
			\includegraphics[width=\columnwidth, clip]{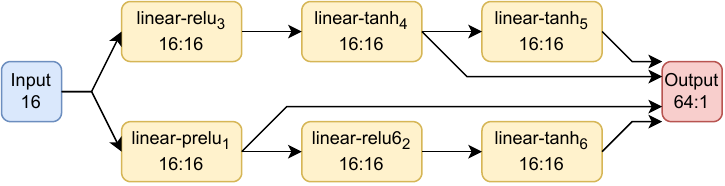}
			\caption{The best neural network found by our reinforcement learning agent. The architecture has $1,716$ trainable parameters, and obtains a median post-quantization F1 Score of 99.884\%. Each node shows the number of input features and output features, separated by a colon.}
			\label{fig:final-network:architecture}
		\end{figure}

		Finally, to fully evaluate the performance of the architecture returned by our reinforcement learning agent, we deploy the best performing architecture on a Google Coral Micro Dev Board~\cite{Google_2022_Dev}, a low-power deep learning accelerator with support for the widely-used TensorFlow Lite deep learning framework~\cite{Abadi_2015_TensorFlow}.
		The architecture selection process happens through the use of the linear utility function from Section~\ref{subsec:methods:nas_agent_training}.
		For each evaluation episode, we take the final architecture, compute the value of the utility function, and select the architecture with the highest utility.
		This architecture is then trained 100 times, each with a  different random initialization, and the trained network with the highest post-quantization F1 score is selected for deployment.
		The architecture obtained a median post-quantization F1 score of 99.884\% on the validation set.
		The lowest post-quantization F1 score obtained on the validation set was 98.917\%.

		The best neural network architecture is showcased in Figure~\ref{fig:final-network:architecture}.
		This architecture has several interesting features.
		First of all, we note that the architecture only contains linear layers with various activation functions.
		Despite the addition of other operations such as convolutions, max pooling and spectral attention, the reinforcement learning agent selected an architecture that consists solely of linear layers.
		The next element of note is the two stream design.
		The agent has essentially designed a network with two parallel information streams of equal depth that are combined at the output node.
		Because of this design, no intermediate node has more than one input, and thus, every intermediate node only has 16 input and 16 output features.
		This reduces the overall parameter count of the neural network, while still retaining depth, allowing for complex decision boundaries with a limited number of trainable parameters.
		We also note that the agent is aggregating features at multiple levels.
		There are branch that connect the input to the output with a path containing 1, 2 and 3 intermediate nodes.
		This is reminiscent of the structure of U-Nets~\cite{Ronneberger_2015_UNet}, or auxiliary towers~\cite{Szegedy_2015_Going} sometimes employed to improve the performance the performance of convolutional neural networks.

		The firmware makes us of the \verb|coralmicro| support library to access the \ac{TPU}.
		Because TensorFlow Lite does not support dynamic batch sizes, images are inferenced in batches of one.
		Data is fed to the Coral Micro Dev Board through an Ethernet-over-\acs{USB} connection.
		The firmware operates on a request-response paradigm and supports two commands.
		The first command is a simple ``PING'' to verify that the network connection between the device and the laptop is operational.
		The second command is an inference request, with a single data sample attached.
		Upon reception of the inference request, the device performs inference, and returns the classification result as an INT8 value representing the probability of the pixel containing fire.

		The version of TensorFlow Lite bundled with the Google Coral Dev Board Micro does not support division operations.
		Our preprocessing code does require division operations, however, since \ac{NDVI}, \ac{NBR} and \ac{AFD} are all instances of the generalized normalized difference index~\cite{Cicala_2018_Landsat}, which requires a division.
		Thus, we opted to execute our preprocessing using the \ac{FPU} present in the ARM M7 core of the NXP i.MX RT1176 microcontroller.
		The data is sent to the device in FP32 format, preprocessing is done executed by the MCU, the data is quantized, and inference of the neural network is executed on the \ac{TPU}.


		\begin{figure*}
			\centering
			\begin{subfigure}[b]{.31\textwidth}
				\centering
				\includegraphics[width=\textwidth, clip]{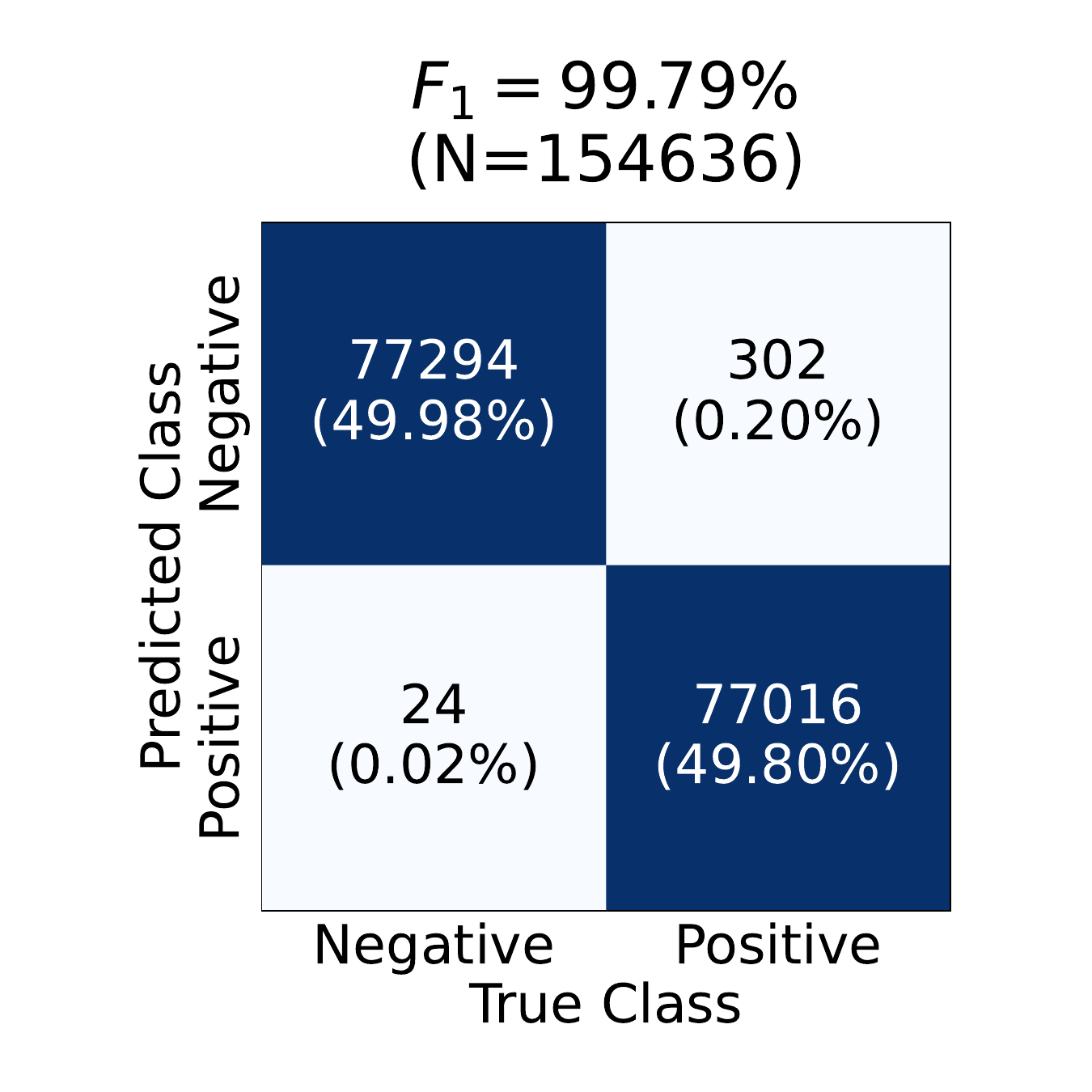}
				\caption{Training set.}
				\label{fig:experiments:deployment:training}
			\end{subfigure}
			\hfill
			\begin{subfigure}[b]{.31\textwidth}
				\centering
				\includegraphics[width=\textwidth, clip]{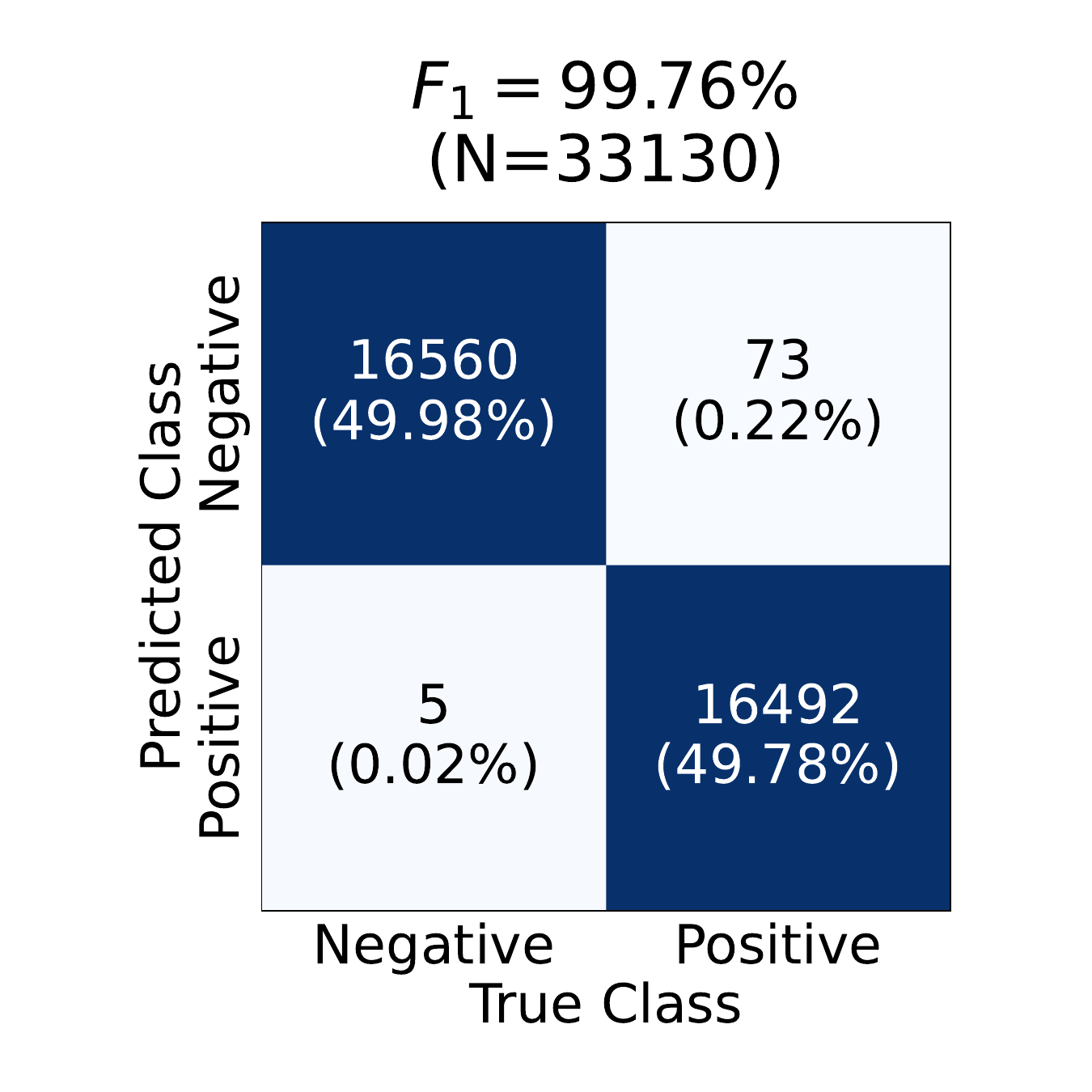}
				\caption{Validation set.}
				\label{fig:experiments:deployment:validation}
			\end{subfigure}
			\hfill
			\begin{subfigure}[b]{.31\textwidth}
				\centering
				\includegraphics[width=\textwidth, clip]{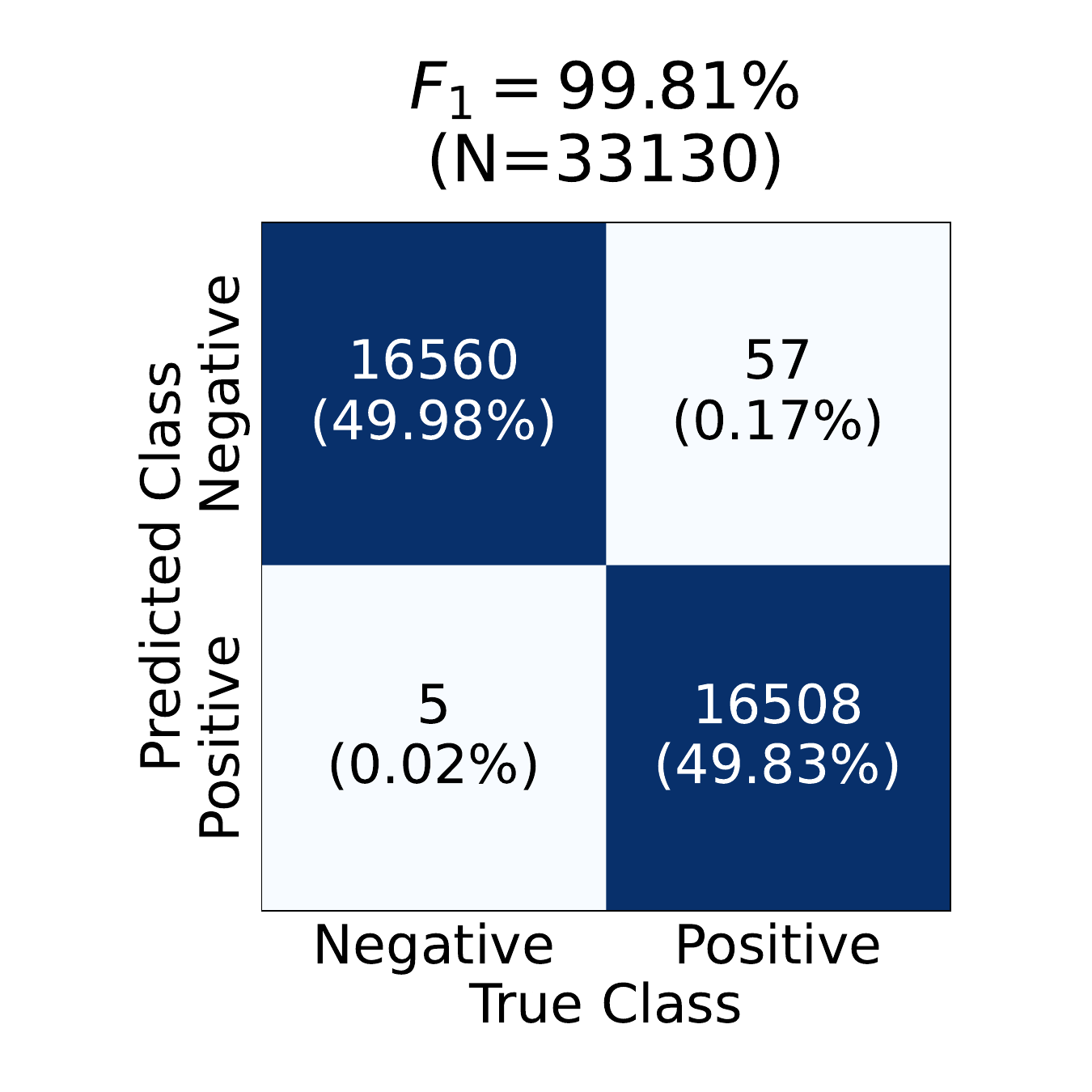}
				\caption{Test set.}
				\label{fig:experiments:deployment:test}
			\end{subfigure}
			\caption{Confusion matrices for the neural network architecture in Figure~\ref{fig:final-network:architecture}.}
			\label{fig:experiments:deployment}
		\end{figure*}

		Figure~\ref{fig:experiments:deployment:training} shows the classification results for the neural network in Figure~\ref{fig:final-network:architecture} for the training set, with results for the validation and test set shown in Figure~\ref{fig:experiments:deployment:validation} and Figure~\ref{fig:experiments:deployment:test}.
		From these confusion matrices, we can conclude that the neural network is a very strong classifier, achieving F1 scores in excess of 99\% on the training, validation and test set.
		We can also see that the network tends to produce false negatives more than it does false positives, with false negatives about $10\times$ more prevalent than false positives.

		When considering only the misclassifications, we see that on each dataset, the neural network classified got at least 2 events perfectly (2 on the training set, 8 on the test set and 10 on the validation set).
		The absolute highest number of misclassifications was made on the ``Australia\_1'' event for the training set, with 168 samples from this set being misclassified (0.2\% of the total samples for ``Australia\_1'' in the training set).
		In relative terms, the ``Greece\_4'' event in the validation set yielded the worst result, with 2.6\% of samples being misclassified (4 out of 154 samples).
		We assess if there's a significant difference in the ranking of the different events between different datasets by computing Kendall's $\tau$ correlation between the ranking of events for different sets.
		The Kendall's $\tau$ correlation between the training and validation set, as well as between the training and testing set is very low, at $-1.05\%$ and $-7.37\%$ respectively.
		Between the validation and testing set we do find a moderate correlation, at $43.16\%$.
		This implies that events that our network did bad at on the validation set, it also did bad at on the test set, which is to be expected under the assumption that there is no overfitting.

		While performing inference on the training set, our firmware averages a round-trip latency of 0.984ms per sample, corresponding to a throughput of 1016.01 samples per second.
		This includes the time it takes to send data from the laptop to the device, and the time it takes to receive the data again.

\begin{figure*}
			\centering
			\begin{subfigure}[b]{.48\textwidth}
				\centering
				\includegraphics[width=\textwidth, clip]{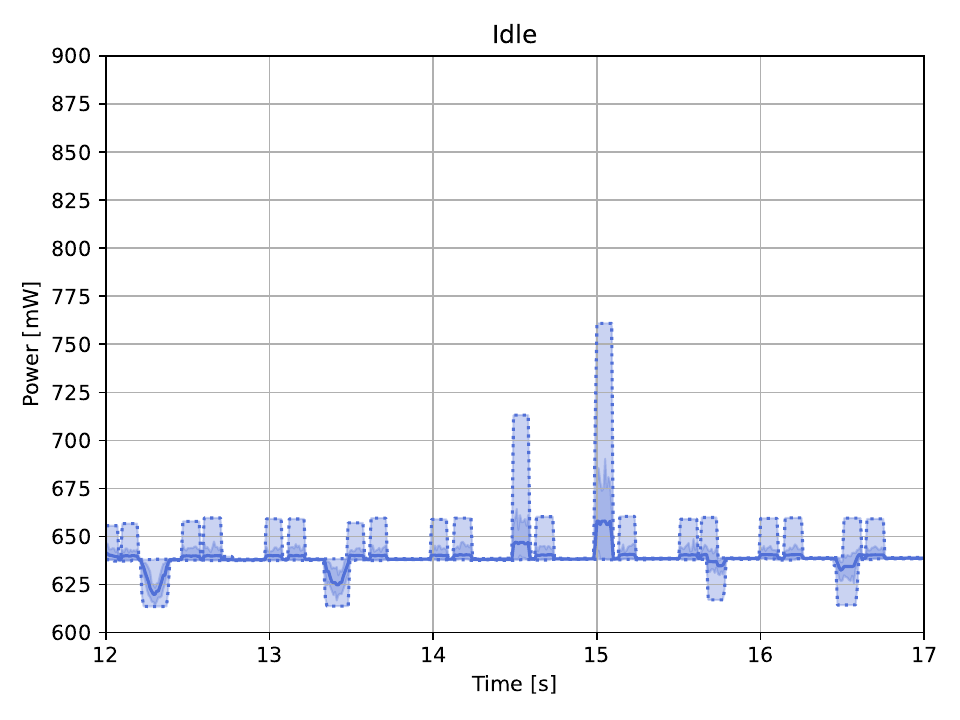}
				\caption{The device is idle, waiting for a command.}
				\label{fig:final-network:joulescope:idle}
			\end{subfigure}
			\hfill
			\begin{subfigure}[b]{.48\textwidth}
				\centering
				\includegraphics[width=\textwidth, clip]{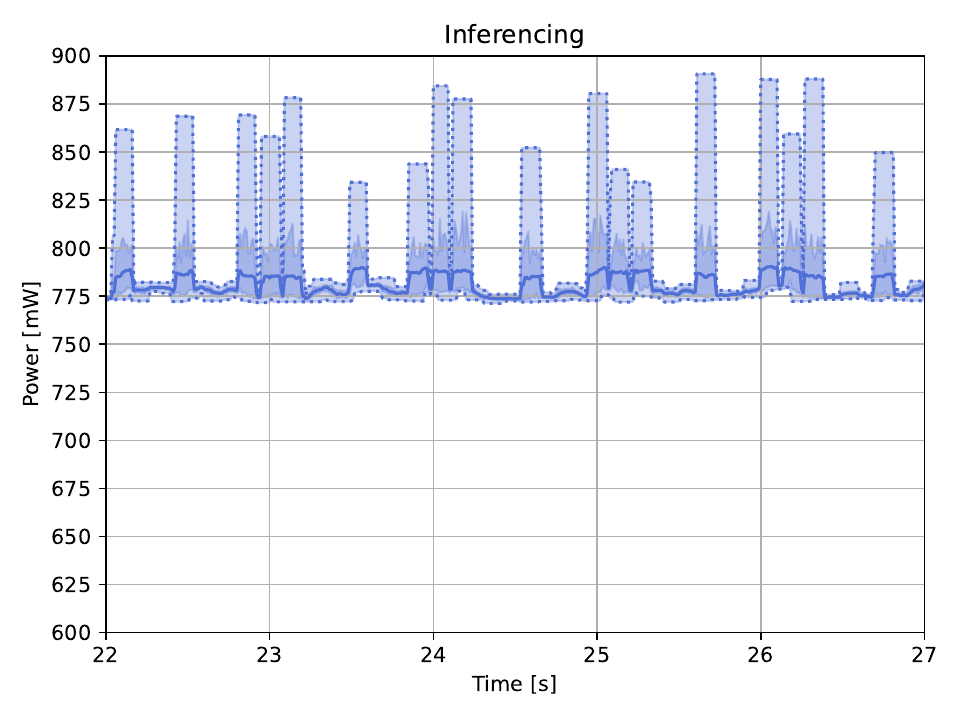}
				\caption{The device is servicing repeated inference requests.}
				\label{fig:final-network:joulescope:inferencing}
			\end{subfigure}
			\caption{A waveform trace of the power consumption of the Google Coral Micro Dev Board while our firmware is running, sampled at 100Hz. Statistics are computed over a 0.1s window. The mean is given as a solid blue line, and the minimum and maximum as a dotted blue line. The area between the minimum and maximum, and the area spanning a 95\% confidence interval around the mean are filled in.}
			\label{fig:final-network:joulescope}
		\end{figure*}

		To analyze the feasibility of deploying this system on a nanosatellite we use a DC Energy Analyzer while the device is operating.
		Our energy analyzer is a JS110 JouleScope~\cite{Jetperch_2024_Joulescope}.
		This should give us insight into the power consumption of the device, both in terms of overall consumed power, as well as peak power consumption.
		Figure~\ref{fig:final-network:joulescope} shows an energy trace while our firmware is running on the Google Coral Micro Dev Board~\cite{Google_2022_Dev}.
		Figure~\ref{fig:final-network:joulescope:idle} shows idle power consumption, while the device is waiting for a command, while Figure~\ref{fig:final-network:joulescope:inferencing} shows the power consumption while the device is performing inference.
		The graph shows that while idle, the device consumes approximately 640mW, which jumps to 780mW average while performing inference.
		The idle power draw for the device is relatively high, which we attribute to several factors.
		First, the firmware we use hasn't been optimized for reducing power consumption.
		Such optimizations include the heavy use of sleep states, and disabling the \ac{TPU} when it is not in use.
		We also note that the communication system used to transfer data to the device (Ethernet-over-\acs{USB}) has a relatively high power draw, with even dedicated \acp{ASIC} drawing in the dozens of milliwatts~\cite{Sung_2024_Ethernet}.
		Despite this seemingly high power consumption, the Dev Board Micro's datasheet~\cite{Google_2022_Dev} references average power peaks of up to 3W which is significantly higher than what we observe.
		Our power consumption being significantly lower is likely explained by the fact that the neural networks we execute are significantly simpler than the complex \acp{CNN} used for computer vision tasks.

\section{Conclusions}
\label{sec:discussion}
	In this publication, we covered the use of a reinforcement learning-based \ac{NAS} strategy to design power-efficient neural networks for the application of detecting active wildfires from satellite imagery.
	We found that the reinforcement learning approach proposed can successfully design neural networks that are accurate at detecting active wildfires, and that use limited computational resources.

	In Section~\ref{subsec:experiments:perf_pred_training} we noted that despite their intrinsic strengths at dealing with graph data, \acp{GNN} performed relatively at the performance prediction task.
	\acp{GNN} are complex models and still an active area of research, we expect that there are many improvements that could be made to the \ac{GNN} architecture we used that would allow us to obtain better predictions.

	Section~\ref{subsec:experiments:perf_pred_training} also showed that while many performance predictors were able to achieve satisfactory performance, there is still room for improvement.
	We expect that the performance of these predictors could be improved through the use of better feature engineering.
	In their paper, \textcite{Kadlecova_2024_Surprisingly} show that the computation of a set of fairly simple features of the underlying architectures can result in a fairly strong feature set to train performance predictors on.
	We hypothesize that using such an improved feature set would likely also lead to stronger predictive performance in our case.

	The reinforcement learning agent originally proposed by \textcite{Cassimon_2024_Scalable} and used in this work proved effective, as shown in Section~\ref{subsec:experiments:nas_agent}.
	While random search and random walks provided a very strong baseline when it comes to reducing the total number of trainable parameters, their ability to design neural networks that can perform accurate classification remains limited.
	The use of both local search and reinforcement learning also revealed some weaknesses in our performance prediction models, in the form of the discovery of adversarial samples.
	Despite this, the final neural network architecture designed by the reinforcement learning agent demonstrated strong performance on the task at hand, and did so within a limited computational budget.

	Finally, in Section~\ref{subsec:experiments:deployment} we deployed the best neural network designed by our reinforcement learning agent onto a Google Coral Dev Board Micro device.
	We find that the models perform well on the target device, and do so within a very limited power envelope, making their deployment on actual smallsat platforms feasible.

\section*{Acknowledgements}
	This research received funding from the Flemish Government (AI Research Program).
	This work was supported by the Research Foundation Flanders (FWO) under Grant Number 1SC8821N.
	This work has been supported by the MOVIQ (Mastering Onboard Vision Intelligence and Quality) project funded by Flanders Innovation \& Entrepreneurship (VLAIO) and Flanders Space (VRI) and has received co-funding from the European Union NextGenerationEU.

\printbibliography

\end{document}